\documentclass[runningheads]{llncs}
\usepackage{graphicx}
\usepackage{tikz}
\usepackage{comment}
\usepackage{amsmath,amssymb} % define this before the line numbering.
\usepackage{color}

\usepackage{booktabs}
\usepackage[pagebackref=true,breaklinks=true,letterpaper=true,colorlinks,bookmarks=false]{hyperref}

\newcommand{\pos}{\mathbf{p}}
\newcommand{\crd}{\mathbf{s}}

\newcommand{\cor}{\mathbf{c}}

\newcommand{\tr}{\mathbf{T}}
\newcommand{\rotm}{\mathbf{R}}
\newcommand{\wei}{\mathbf{w}}
\newcommand{\img}{\emph{I}}

\newcommand{\be}{\mathbf{t}}
\newcommand{\bI}{\mathbf{I}}
\newcommand{\bbX}{\mathbf{\bar{X}}}

\newcommand{\bX}{\mathbf{X}}

\newcommand{\lossh}{\mathcal{L}}

\begin{document}
% \linenumbers
\pagestyle{headings}
\mainmatter
\def\ECCVSubNumber{3498}  % Insert your submission number here

\title{SC-wLS: Towards Interpretable Feed-forward Camera Re-localization}

%% INITIAL SUBMISSION 
%%\begin{comment}
%\titlerunning{ECCV-22 submission ID \ECCVSubNumber} 
%\authorrunning{ECCV-22 submission ID \ECCVSubNumber} 
%\author{Anonymous ECCV submission}
%\institute{Paper ID \ECCVSubNumber}
%%\end{comment}
%******************

\makeatletter
\newcommand{\printfnsymbol}[1]{%
	\textsuperscript{\@fnsymbol{#1}}%
}
\makeatother

%\begin{comment}
\titlerunning{SC-wLS: Towards Interpretable Feed-forward Camera Re-localization}
\author{Xin Wu\inst{1,2}\thanks{equal contribution} \and
Hao Zhao\inst{1,3}\printfnsymbol{1} \and
Shunkai Li\inst{4} \and
Yingdian Cao\inst{1,2} \and
Hongbin Zha\inst{1,2} }
\authorrunning{X. Wu et al.}
\institute{Key Laboratory of Machine Perception (MOE), School of AI, Peking University \and
PKU-SenseTime Machine Vision Joint Lab \and
Intel Labs China \and Kuaishou Technology \\
\email{\{wuxin1998,zhao-hao,lishunkai,yingdianc\}@pku.edu.cn, zha@cis.pku.edu.cn} \\
\url{https://github.com/XinWu98/SC-wLS}  }
%\end{comment}
%******************
\maketitle

\begin{abstract}
   Visual re-localization aims to recover camera poses in a known environment, which is vital for applications like robotics or augmented reality. Feed-forward absolute camera pose regression methods directly output poses by a network, but suffer from low accuracy. Meanwhile, scene coordinate based methods are accurate, but need iterative RANSAC post-processing, which brings challenges to efficient end-to-end training and inference. In order to have the best of both worlds, we propose a feed-forward method termed SC-wLS that exploits all \textbf{s}cene \textbf{c}oordinate estimates for \textbf{w}eighted \textbf{l}east \textbf{s}quares pose regression. This differentiable formulation exploits a weight network imposed on 2D-3D correspondences, and requires pose supervision only. Qualitative results demonstrate the interpretability of learned weights. Evaluations on 7Scenes and Cambridge datasets show significantly promoted performance when compared with former feed-forward counterparts. Moreover, our SC-wLS method enables a new capability: self-supervised test-time adaptation on the weight network. Codes and models are publicly available.
   %at \url{https://github.com/Alice51129/SC-wLS}.
   
% The abstract should summarize the contents of the paper. LNCS guidelines
% indicate it should be at least 70 and at most 150 words. It should be set in 9-point
% font size and should be inset 1.0~cm from the right and left margins.
% \dots
\end{abstract}

\section{Introduction}
Visual re-localization \cite{cao2013graph,hays2008im2gps,sattler2016efficient,shotton2013scene} determines the global 6-DoF poses (\textit{i.e.}, position and orientation) of query RGB images in a known environment. It is a fundamental computer vision problem and has many applications in robotics and augmented reality. Recently there is a trend to incorporate deep neural networks into various 3D vision tasks, and use differentiable formulations that optimize losses of interest to learn result-oriented intermediate representation. Following this trend, many learning-based absolute pose regression (APR) methods \cite{kendall2015posenet,brahmbhatt2018geometry} have been proposed for camera re-localization, which only need a single feed-forward pass to recover poses.
% They impose end-to-end training and inference, and are robust in challenging environments.
However, they treat the neural network as a black box and suffer from low accuracy \cite{sattler2019understanding}. On the other hand, scene coordinate based methods learn pixel-wise 3D scene coordinates from RGB images and solve camera poses using 2D-3D correspondences by Perspective-n-Point (PnP) \cite{li2012robust}. In order to handle outliers in estimated scene coordinates, the random sample consensus (RANSAC) \cite{choi1997performance} algorithm is usually used for robust fitting. Compared to the feed-forward APR paradigm, scene coordinate based methods achieve state-of-the-art performance on public camera re-localization datasets. However, RANSAC-based post-processing is an iterative procedure conducted on CPUs, which brings engineering challenges for efficient end-to-end training and inference \cite{brachmann2017dsac,brachmann2021visual}.

\begin{figure}[tb]
	\begin{center}
		\includegraphics[scale=0.3]{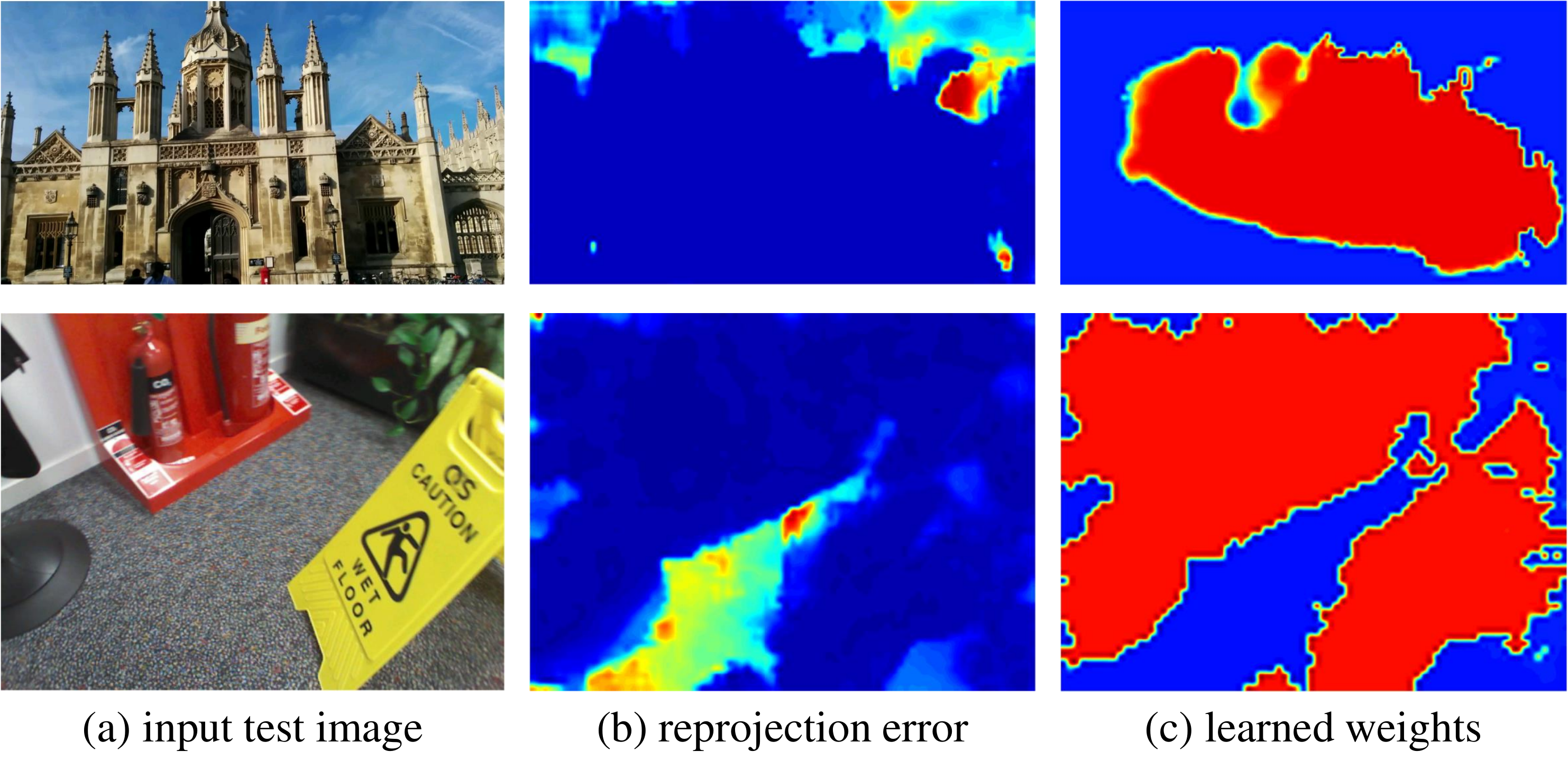}
	\end{center}
	\vspace{-10pt}
	\caption{For input images (a), our network firstly regresses their scene coordinates, then predicts correspondence-wise weights (c). With these weights, we can use all 2D-3D correspondences for end-to-end differentiable least squares pose estimation. We use re-projeciton errors (b) to illustrate scene coordinate quality. Our weights select high-quality scene coordinates. A higher color temperature represents a higher value.}
	\label{fig:teaser}
\end{figure}

In order to get the best of both worlds, we develop a new feed-forward method based upon the state-of-the-art (SOTA) pipeline DSAC* \cite{brachmann2021visual}, thus enjoying the strong representation power of scene coordinates. We propose an alternative option to RANSAC that exploits all 3D \textbf{s}cene \textbf{c}oordinate estimates for \textbf{w}eighted \textbf{l}east \textbf{s}quares pose regression (SC-wLS). The key to SC-wLS is a weight network that treats 2D-3D correspondences as 5D point clouds and learns weights that capture geometric patterns in this 5D space, with only pose supervision. Our learned weights can be used to interpret how much each scene coordinate contributes to the least squares solver.

Our SC-wLS estimates poses using only tensor operators on GPUs, which is similar to APR methods due to the feed-forward nature but out-performs APR methods due to the usage of scene coordinates. Furthermore, we show that a self-supervised test-time adaptation step that updates the weight network can lead to further performance improvements. This is potentially useful in scenarios like a robot vacuum adapts to specific rooms during standby time. Although we focus on comparisons with APR methods, we also equip SC-wLS with the LM-Refine post-processing module provided in DSAC* \cite{brachmann2021visual} to explore the limits of SC-wLS, and show that it out-performs SOTA on the outdoor dataset Cambridge.%On the other hand, due to the strong representation power of scene coordinates, SC-wLS is much more accurate than APR methods.

Our major contributions can be summarized as follows: (1) We propose a new feed-forward camera re-localization method, termed SC-wLS, that learns interpretable scene coordinate quality weights (as in Fig.~\ref{fig:teaser}) for weighted least squares pose estimation, with only pose supervision. (2) Our method combines the advantages of two paradigms. It exploits learnt 2D-3D correspondences while still allows efficient end-to-end training and feed-forward inference in a principled manner. As a result, we achieve significantly better results than APR methods. (3) Our SC-wLS formulation allows test-time adaptation via self-supervised fine-tuning of the weight network with the photometric loss.

\section{Related works}

\textbf{Camera re-localization.} In the following, we discuss camera re-localization methods from the perspective of map representation. 

Representing image databases with global descriptors like thumbnails \cite{hays2008im2gps}, 
% GIST \cite{oliva2001modeling}
BoW \cite{sivic2003video}, or learned features \cite{arandjelovic2016netvlad} is a natural choice for camera re-localization. By retrieving poses of similar images, localization can be done in the extremely large scale \cite{hays2008im2gps,schindler2007city}.
%  By retrieving the poses of similar images, localization can be done in large scale \cite{schindler2007city}. Recently, improved techniques on both descriptors \cite{torii201524} and databases \cite{cao2013graph} have been proposed. 
%  CNN-based absolute pose regression 
Meanwhile, CNN-based absolute pose regression methods \cite{kendall2015posenet,walch2017image,brahmbhatt2018geometry,xue2020learning} belong to this category, since their final-layer embeddings are also learned global descriptors. They regress camera poses from single images in an end-to-end manner, and recent work primarily focuses on sequential inputs \cite{xue2019local} and network structure enhancement \cite{xue2020learning,shavit2021learning,ding2019camnet}. Although the accuracies of this line of methods are generally low due to intrinsic limitations \cite{sattler2019understanding}, they are usually compact and fast, enabling pose estimation in a single feed-forward pass.
% they enable pose estimation in a single forward pass, which are usually fast and light. 

% they are useful in producing initial pose estimations \cite{taira2018inloc}\cite{sarlin21pixloc}. 

Maps can also be represented by 3D point cloud \cite{wu2013towards} with associated 2D descriptors \cite{lowe2004distinctive} via SfM tools \cite{schonberger2016structure}. Given a query image, feature matching establishes sparse 2D-3D correspondences and yields very accurate camera poses with RANSAC-PnP pose optimization \cite{zhang2006image,sattler2016efficient}. The success of these methods heavily depends on the discriminativeness of features and the robustness of matching strategies. Inspired by feature based pipelines, scene coordinate regression learns a 2D-3D correspondence for each pixel, instead of using feature extraction and matching separately. The map is implicitly encoded into network parameters. \cite{mair2009efficient} demonstrates impressive localization performance using stereo initialization and sequence input.
% As for their performance comparision, 
Recently, \cite{brachmann2021limits} shows that the algorithm used to create pseudo ground truth has a significant impact on the relative ranking of above methods.

Apart from random forest based methods using RGB-D inputs \cite{shotton2013scene,valentin2015exploiting,meng2018exploiting}, scene coordinate regression on RGB images is seeing steady progress  \cite{brachmann2017dsac,Brachmann2018LearningLI,Brachmann2019NeuralGuidedRL,brachmann2021visual,zhou2020kfnet}. 
This line of work lays the foundation for our research. 
In this scheme, predicted scene coordinates are noisy due to single-view ambiguity and domain gap during inference. As such, \cite{brachmann2017dsac,Brachmann2018LearningLI} use RANSAC and non-linear optimization to deal with outliers, and NG-RANSAC \cite{Brachmann2019NeuralGuidedRL} learns correspondence-wise weights to guide RANSAC sampling.
% pose estimation purely relys on the output of scene coordinates regression networks, which can be noisy and unrecognizable due to the single-view training process and the blackbox property of neural networks.
% To deal with outliers, \cite{brachmann2017dsac}\cite{Brachmann2018LearningLI} use the RANSAC algorithm, sampling iteratively to select the best hypothesis. Additionally, NG-RANSAC \cite{Brachmann2019NeuralGuidedRL} proposes to learn a neural weight with differentiable RANSAC \cite{brachmann2017dsac}to guide this sampling process. 
%However, these robust estimators need iterative sampling which brings engineering challenges for efficient training and inference, and
\cite{Brachmann2019NeuralGuidedRL} conditions weights on RGB images, whose statistics is often influenced by factors
like lighting, weather or even exposure time. Object pose and room layout estimation \cite{lepetit2005monocular,zhong2020seeing,hedau2009recovering,hirzer2020smart,zhao2017physics,yan20203d} can also be addressed with similar representations \cite{brachmann2014learning,wang2019normalized}.

%By contrast, we propose SC-wLS to weight scene coordinates by a fully differentiable neural network. It relies purely on geometric information and exploits global patterns in one shot. As such, our learnt weights can align with fine details for both indoor and outdoor environments, effectively give reasonable pose estimation in a feed-forward pass. The decoupling of weight regression also enables a self-supervised adaptation at test-time.
% guiding later pose optimization.

\noindent \textbf{Differentiable optimization.}
% In 3D geometry vision field, to make traditional optimization modules compatible with prevalent deep neural networks, some research focuses on formulate geometry optimization in an end-to-end trainable fashion. As such, the framework is fully differentiable and able to learn result-oriented feature representation.
% Deep neural networks have shown great power in vision tasks, 
% % are good at learning discriminative models, 
% yet there are still many 3D geometric tasks including optimization modules, which are not differentiable and hinder the learning of result-oriented feature. To tackle this problem, some research focuses on formulate geometry optimization in an end-to-end trainable fashion.
To enhance their compatibility with deep neural networks and learn result-oriented feature representation, some recent works focus on re-formulating geometric optimization techniques into an end-to-end trainable fashion, for various 3D vision tasks. \cite{ionescu2015matrix} proposes several standard differentiable optimization layers. \cite{Yi2018LearningTF,ranftl2018deep} propose to estimate fundamental/essential matrix via solving weighted least squares problems with spectral layers.
% Solving least squares problems with spectral layers has shown successful applications in fundamental/essential matrix estimation \cite{Yi2018LearningTF}\cite{ranftl2018deep} or primitive fitting in point clouds \cite{li2019supervised}.
\cite{Dang2020EigendecompositionFreeTO} further shows that the eigen-value switching problem when solving least squares can be avoided by minimizing linear system residuals.
% minimizing linear system residuals can avoid the eigen-value switching problem. 
\cite{gould2021deep} develops generic black-box differentiable optimization techniques with implicit declarative nodes.

\begin{figure*}[tb]
	\begin{center}
\includegraphics[width=0.9\linewidth]{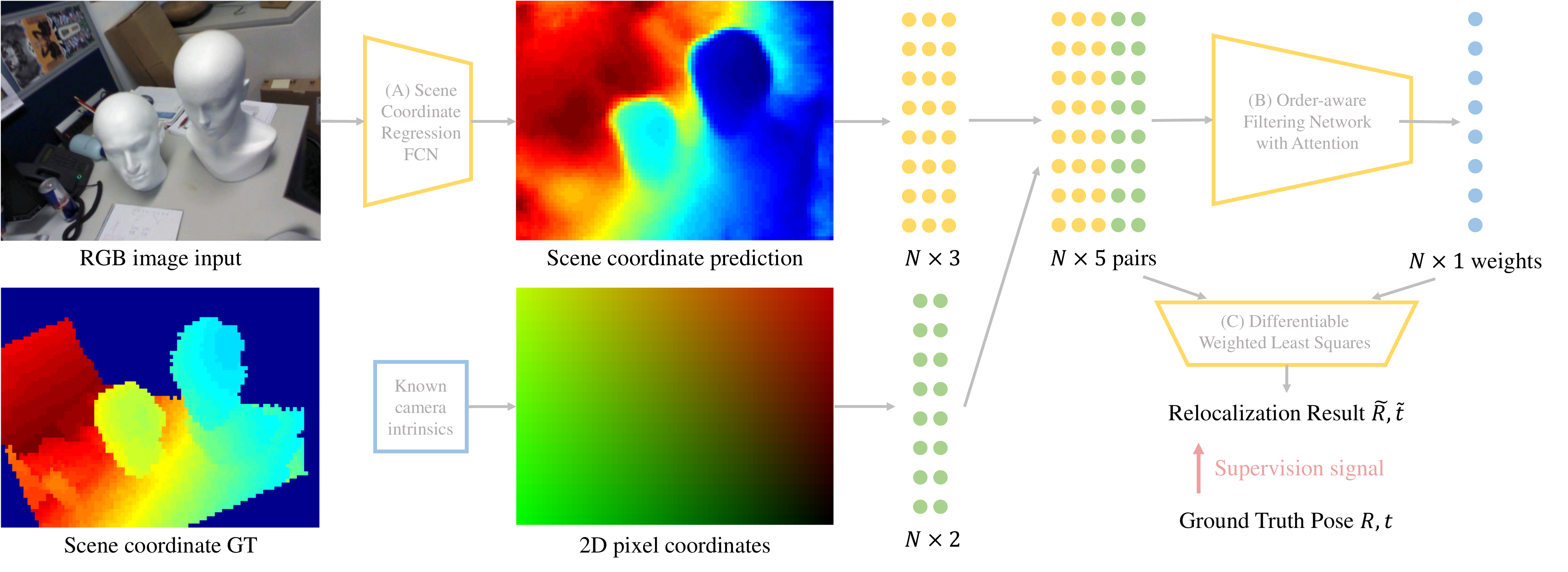}
	\end{center}
	\vspace{-10pt}
	\caption{The overview of SC-wLS. Firstly, a fully convolutional network (A) regresses pixel-wise scene coordinates from an input RGB image. Scene coordinate predictions are flattened to the shape of $N\times3$, with $N$ being pixel count. We concatenate it with normalized $N\times2$ 2D pixel coordinates, forming $N\times5$ correspondence inputs. The correspondences are fed into the weight learning network (B), producing $N\times1$ weights indicating scene coordinate quality. The architecture of B is an order-aware filtering network \cite{zhang2019learning} with graph attention modules \cite{sarlin2020superglue}. Thirdly, correspondences and weights are sent into a differentiable weighted least squares layer (C), directly outputing camera poses. The scene coordinate ground truth is not used during training. }
	\label{fig:arch}
	\vspace{0pt}
\end{figure*}

\section{Method}

Given an RGB image $I$, we aim to find an estimate of the absolute camera pose consisting of a 3D translation and a 3D rotation, in the world coordinate system. Towards this goal, we exploit the \textbf{s}cene \textbf{c}oordinate representation. Specifically, for each pixel $i$ with position $\pos_i$ in an image, we predict the corresponding 3D scene coordinate $\crd_i$. As illustrated in Fig.~\ref{fig:arch}, we propose an end-to-end trainable deep network that directly calculates global camera poses via \textbf{w}eighted \textbf{l}east \textbf{s}quares. The method is named as \textbf{SC-wLS}. Fig.~\ref{fig:arch}-A is a standard fully convolutional network for scene coordinate regression, as used in former works \cite{brachmann2021visual}. Our innovation lies in Fig.~\ref{fig:arch}-B/C, as elaborated below. %It exploits the geometric consistency between 2D pixels and learned 3D scene coordinates.

\subsection{Formulation}

Given learned 3D scene coordinates and corresponding 2D pixel positions, our goal is to determine the absolute poses of calibrated images taking all correspondences into account. This would inevitably include outliers, and we need to give them proper weights. Ideally, if all outliers are rejected by zero weights, calculating the absolute pose can be formulated as a linear least squares problem. Inspired by \cite{Yi2018LearningTF} (which solves an essential matrix problem instead), we use all of the $N$ 2D-3D correspondences as input and predict $N$ respective weights $\wei_{i}$ using a neural network (Fig.~\ref{fig:arch}-B). $\wei_{i}$ indicates the uncertainty of each scene coordinate prediction. As such the ideal least squares problem is turned into a weighted version for pose recovery.

Specifically, the input correspondence $\cor_{i}$ to Fig.~\ref{fig:arch}-B is
\begin{equation}
\cor_{i} = [x_{i}, y_{i}, z_{i}, u_{i}, v_{i}]
\end{equation}
where $x_{i}, y_{i}, z_{i}$ are the three components of the scene coordinate $\crd_{i}$, and $u_i$, $v_i$ denote the corresponding pixel position. $u_i$, $v_i$ are generated by normalizing $\pos_{i}$ with the known camera intrinsic matrix. 
The absolute pose is written as a transformation matrix $\tr \in \mathbb{R}^{3\times4} $. It projects scene coordinates to the camera plane as below:
\begin{equation} \label{eq:transform}
\begin{bmatrix}
u_{i}\\
v_{i}\\
1
\end{bmatrix}\!{=}
 \tr 
\begin{bmatrix}
x_{i}\\
y_{i}\\
z_{i}\\
1
\end{bmatrix}\!{=}\!
\begin{bmatrix}
 p_{1} &p_{2} &p_{3} &p_{4} \\ 
 p_{5} &p_{6} &p_{7} &p_{8} \\ 
 p_{9} &p_{10} &p_{11} &p_{12} \\
\end{bmatrix}
\begin{bmatrix}
x_{i}\\
y_{i}\\
z_{i}\\
1
\end{bmatrix}\;
\end{equation}
When $N > 6$, 
the transformation matrix $\tr$ can be recovered by Direct Linear Transform (DLT) \cite{Hartley2004MultipleVG}, which converts Eq.~\ref{eq:transform} into a linear system:

\begin{equation} \label{eq:LS}
	\mathbf X \rm Vec(\tr) = 0
\end{equation}

$\rm Vec(\tr)$ is the vectorized $\mathbf T$. $\mathbf X$ is a $\mathbb{R}^{2N \times 12} $ matrix whose $2i-1$ and $2i$ rows $\mathbf X^{(2i{-}1)}$ , $\mathbf X^{(2i)}$ are as follows:

\small
\begin{equation}
	\begin{bmatrix}
		x_{i},  y_{i}, z_{i}, 1, \; 0, \; 0, \; 0, \; 0,  -u_{i}x_{i},  -u_{i}y_{i},  -u_{i}z_{i},  -u_{i} \\
		0, \; 0, \; 0, \; 0, \; x_{i}, y_{i}, z_{i}, 1,  -v_{i}x_{i},  -v_{i}y_{i},  -v_{i}z_{i},  -v_{i}
	\end{bmatrix}
	\label{eq:datamat_pnp}
\end{equation} \normalsize

As such, pose estimation is formulated as a least squares problem. $\rm Vec(\tr)$ can be recovered by finding the eigenvector associated to the smallest eigenvalue of $\mathbf X^\top\mathbf X$.

Note that in SC-wLS, each correspondence contributes differently according to $\wei_{i}$, so $\mathbf X^{\top}\mathbf X$ can be rewritten as $\mathbf X^{\top}\rm diag(\wei)\mathbf X$
and $\rm Vec(\tr)$ still corresponds to its smallest eigenvector. As the rotation matrix $\rotm$ needs to be orthogonal and has determinant $1$, we further refine the DLT results by the generalized Procrustes algorithm \cite{schonemann1966generalized}, which is also differentiable. More details about this post-processing step can be found in the supplementary material.

\subsection{Network design}

Then we describe the architecture of Fig.~\ref{fig:arch}-B. We treat the set of correspondences $\{\cor_{i}\}$ as unordered 5-dimensional point clouds and resort to a PointNet-like architecture \cite{zhang2019learning} dubbed OANet. This hierarchical network design consists of multiple components including DiffPool layers, Order-Aware filtering blocks and DiffUnpool layers. It can guarantee the permutation-invariant property of input correspondences. Our improved version enhances the Order-Aware filtering block with self-attention. 

\begin{figure}[tb]
	\begin{center}
		\includegraphics[scale=0.35]{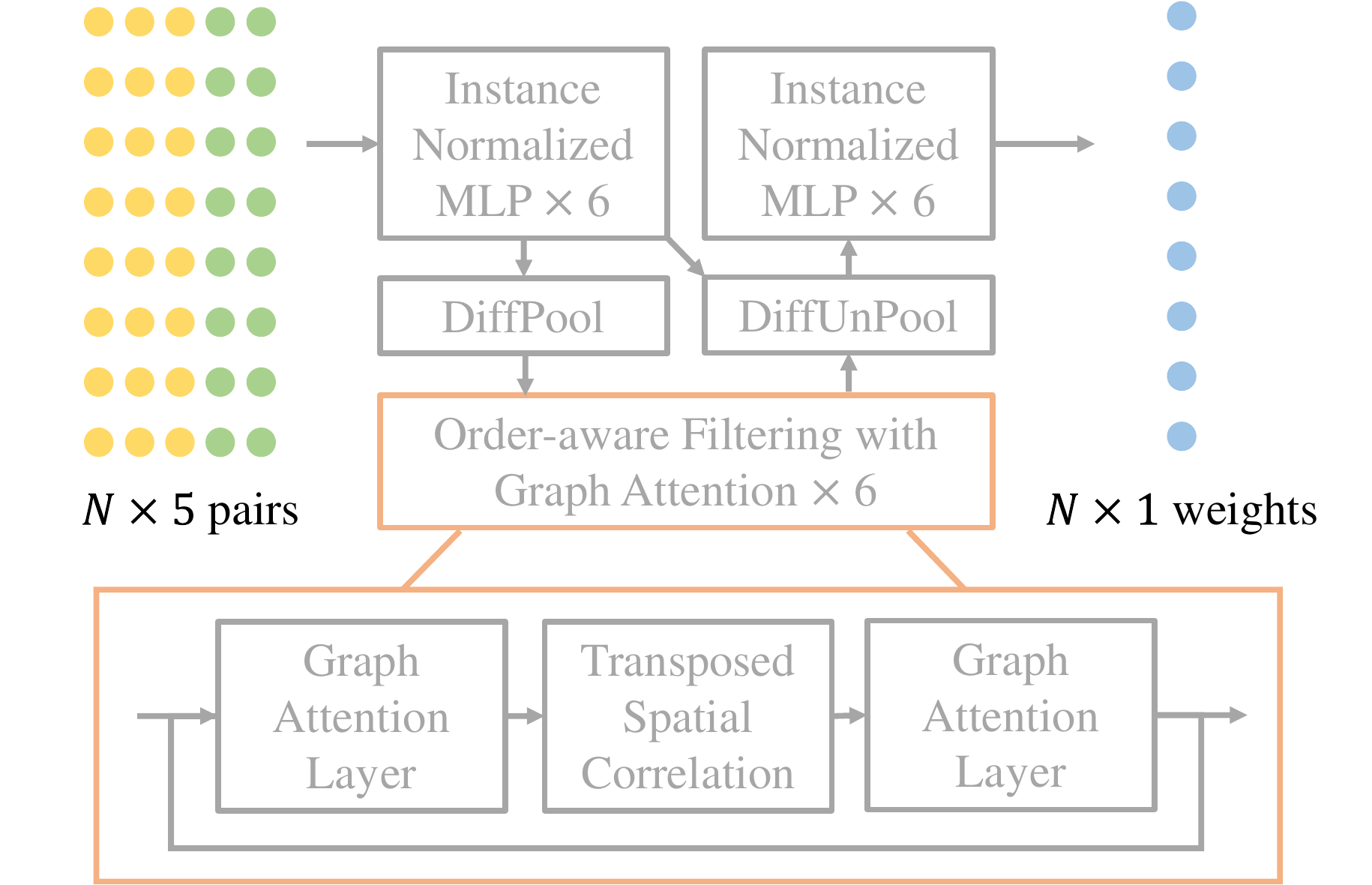}
	\end{center}
	\vspace{-10pt}
	\caption{The architecture of Fig.~\ref{fig:arch}-B, which is inherited from OANet \cite{zhang2019learning}. Here we use two self-attention graph layers from \cite{sarlin2020superglue} while in the original OANet they are instance normazlied MLPs.}
	\label{fig:oanet}
\end{figure}

The network is illustrated in Fig.~\ref{fig:oanet}. Specifically speaking, the DiffPool layer firstly clusters inputs into a particularly learned canonical order. Then the clusters are spatially correlated to retrieve the permutation-invariant context and finally recovered back to original size through the DiffUnpool operator. %These operations effectively overcome the drawbacks of former PointNet-like architectures \cite{qi2017pointnet} \cite{qi2017pointnet++}. \cite{qi2017pointnet} suffers from the lack of context and \cite{qi2017pointnet++} is restricted by hand-crafted neighborhood selection. Besides, this design can also process different number of correspondences, which enables a flexible data augmentation. %In practice, we pool correspondences into 500 clusters for memory and computation efficiency. 
Inspired by the success of transformer \cite{vaswani2017attention} and its extension in 3D vision \cite{sarlin2020superglue}, we propose to introduce the self-attention mechanism into OANet to better reason about the underlying relationship between correspondences. As shown in Fig.~\ref{fig:oanet}, there are two instance normalized MLP modules before and after the transposed spatial correlation module, in original OANet. We replace them with attention-based message passing modules, which can exploit the spatial and contextual information simultaneously. 

The clusters can be regarded as nodes $\mathcal V = \{ v_{i} \}$ in a graph $\mathcal{G}$ and the edges $\mathcal E = \{ e_{i} \}$ are constructed between every two nodes. Since the number of clusters is significantly less than that of original inputs, calculating self-attention on these fully connected weighted edges would be tractable, in terms of computation speed and memory usage. Let ${}^{(\rm in)}\mathbf{f}_i$ be the intermediate representation for node $v_{i}$ at input, and the self-attention operator can be described as:
 \begin{equation}
    \label{eqn:message-passing}
        {}^{(\rm out)}\mathbf{f}_i =
        {}^{(\rm in)}\mathbf{f}_i
        + \text{MLP}\left( \left[{}^{(\rm in)}\mathbf{f}_i\,||\,\mathbf{m}_{\mathcal{E}\rightarrow i}\right]\right),
    \end{equation}
where $\mathbf{m}_{\mathcal E \rightarrow i}$ is the message aggregated from all other nodes $\{j: (i, j)\in\mathcal{E}\}$ using self-attention described in \cite{sarlin2020superglue}, and $[\cdot\,||\,\cdot]$ denotes concatenation.  

Finally, we apply two sequential activation functions (a ReLU followed by a Tanh) upon the outputs of Fig.~\ref{fig:oanet} to get the weights $\wei_{i}$ in the range $[0, 1)$, for indoor scenes. We use the log-sigmoid activation for outdoor environments following \cite{Brachmann2019NeuralGuidedRL}, for more stable gradient back-propagation.

\subsection{Loss functions}
\label{sec:loss func}

%\textbf{Add}
%Generally, scene coordinates regression methods are supervised by groud-truth scene coordinates, which are generated by reconstruction and easily fail in texture-less areas or repeating structures. In the real world camera relocalization applications, running such reconstruction pipeline needs heavy computation. To this end, our whole framework is solely supervised by ground-truth poses without accessing any 3D models.

Generating ground-truth scene coordinates for supervision is time-consuming for real-world applications. To this end, our whole framework is solely supervised by ground-truth poses without accessing any 3D models. Successfully training the network with only pose supervision requires three stages using different loss functions. Before describing the training protocol, we first define losses here.
%To predict scene coordinates and corresponding weights with only poses as supervision, different losses are leveraged here, including re-projection loss, classification loss, and regression loss.

\textbf{Re-projection loss} is defined as follows:

\begin{equation}
	\label{eq:proj}
	r_i = ||\mathbf{K}\tr^{-1}\crd_i - \pos_i||_2,
\end{equation} 

\begin{equation}
	\label{eq:lossp}
	\lossh_{p}^{(i)}= 
	\begin{cases}
		r_i   & \text{if } \crd_i \in \rm V\\
		||\bar{\crd}_i - \crd_i||_1               & \text{otherwise}.
	\end{cases}
\end{equation}

\begin{equation}
	\label{eq:lossp2}
	\lossh_{p}= \frac{1}{N}\sum_{i=1}^{N}\lossh_{p}^{(i)}
\end{equation} 

$\mathbf{K}$ is the known camera intrinsic matrix, and $\rm V$ is a set of predicted $\crd_i$ that meet some specific validity constraints following DSAC* \cite{brachmann2021visual}. If $\crd_i \in \rm V$, we use the re-projection error $r_i$ in Eq.~\ref{eq:proj}. Otherwise, we generate a heuristic $\bar{\crd}_i$ assuming a depth of $10$ meters. This hyper-parameter is inherited from \cite{brachmann2021visual}.

%To learn reasonable scene coordinate weights for DLT pose estimation, our final end-to-end objective function takes two terms into account, following \cite{Yi2018LearningTF}:

%\begin{equation}
%\label{eq:loss_all}
%\lossh = \lossh_{c} + \gamma \lossh_{r}
%\end{equation}

%where $\lossh_{c}$ is a classification loss and $\lossh_{r}$ is a regression loss, while $\gamma$ balances them. 

% \vspace{-5mm}
\textbf{Classification loss $\lossh_{c}$ } Without knowing the ground truth for scene coordinate quality weights, we utilize the re-projection error $r_i$ for weak supervision:

\begin{equation}
\label{eq:label}
l_{i}= \begin{cases}
0, & \text {if $r_i > \tau$} \\
1, &\text{otherwise}
\end{cases} 
\end{equation} 
\begin{equation}
  %% \loss_{x}(\Phi,\bx_k)= -\frac{1}{N}\sum_{i=1}^{N}\gamma_{k}^i\log\left(\sigmoid\left(y_k^io_k^i\right)\right) \; ,
  \lossh_{c}= \frac{1}{N}\sum_{i=1}^{N}H\left(l_{i}, \wei_{i}\right) \; ,
  \label{eq:classification_loss}
\end{equation}
where $H$ is the binary cross-entropy function. $\tau$ is empirically set to $1$ pixel to reject outliers. Ablation studies about $\lossh_{c}$ can be found in the supplementary.

% \vspace{-5mm}
\textbf{Regression loss $\lossh_{r}$ } Since our SC-wLS method is fully differentiable using the eigen-decomposition (ED) technique \cite{ionescu2015matrix}, imposing $ L_{2}$ or other losses on the transform matrix $\tr$ is straighfoward. However, as illustrated in \cite{Dang2020EigendecompositionFreeTO}, the eigen-decomposition operations would result in gradient instabilities due to the eigen vector switching phenomenon. Thus we utilize the eigen-decomposition free loss \cite{Dang2020EigendecompositionFreeTO} to constrain the pose output, which avoids explicitly performing ED so as to guarantee convergence:
  \begin{equation}
    \begin{aligned}
      \lossh_{r} =
     \be^\top \bX^\top \rm diag(\wei) \bX \be\ + \alpha e^{-\beta tr(\bbX^\top \rm diag(\wei) \bbX)} \,
    \end{aligned}
    \label{eq:ell_comb}
  \end{equation}
  where $\be$ is the flattened ground-truth pose, $\bbX {=} \bX(\bI - \be\be^\top)$, $\alpha$ and $\beta$ are positive scalars. Two terms in Eq.~\ref{eq:ell_comb} 
  serve as different roles. The former tends to minimize the error of pose estimation while the latter tries to alleviate the impact of null space. The trace value in the second term can change up to thousands for different batches, thus the hyper-parameter $\beta$ should be set to balance this effect. %More details about $\alpha$ and $\beta$ can be found in \cite{Dang2020EigendecompositionFreeTO}.

  \subsection{Training protocol}
  \label{sec:training protocol}
  
We propose a three-stage protocol with
different objective functions to train our network Fig.~\ref{fig:arch}. Note that ground truth poses are known in all three stages.
%As mentioned above, for better utility in real-world applications, we train SC-wLS using only absolute poses instead of 3D models.

% Our pipeline can be trained in an end-to-end fashion using
% pairs of \rgb images and ground truth poses, but doing
% so from scratch will fail as the system quickly reaches
% a local minimum.We propose a new 3-step training schema with
% different objective functions in each step. Depending on whether a 3D scene model is available or not, we use rendered
% or approximate scene coordinates to initialize the network
% in the first step.(Impractical and laborious to obtain 3D scene model by SfM)
% (When a dense  3D scan of an environment is unavailable, SfM tools like
% VisualSfM [9] or COLMAP [10] offer workable solutions to
% create a (sparse) 3D model from a collection of RGB images,
% e.g. from the training set of a scene. However, for some
% environments, particularly indoors, a SfM reconstruction
% might fail due to texture-less areas or repeating structures.
% Also, despite SfM tools having matured significantly over
% many years since the introduction of Bundler [8] they
% still represent expert tools with their own set of hyperparameters
% to be tuned. Therefore, it might be attractive to train a camera re-localization system from RGB images and ground truth poses alone, without resorting to an SfM tool
% for pre-processing.)Training steps two and three improve
% the accuracy of the system which is crucial when no 3D
% model was provided for initialization.

% \noindent \textbf{Scene coordinate initialization.}
\noindent \textbf{Scene coordinate initialization.}
Firstly, we train our scene coordinate regression network (Fig.~\ref{fig:arch}-A) by optimizing the re-projection error $\lossh_{p}$ in Eq.~\ref{eq:lossp2}, using ground truth poses. This step gives us reasonable initial scene coordinates, which are critical for the convergence of $\lossh_{c}$ and $\lossh_{r}$. %By this step, we can get comparable scene coordinate results with the one trained with both \rgb images and 3D model.

\noindent \textbf{Weight initialization.} Secondly, we exploit $\lossh = \lossh_{c} + \gamma \lossh_{r}$ to train Fig.~\ref{fig:arch}-B, where $\gamma$ is a balancing weight. The classification loss $\lossh_{c}$ is used to reject outliers. The regression loss $\lossh_{r}$ is used to constrain poses, making a compromise between the accuracy of estimated weights and how much information is reserved in the null-space. $\lossh_{c}$ is important to the stable convergence of this stage.%Thus the parameters $\alpha$ and $\beta$ in Eq.\ref{eq:ell_comb} should be properly selected. We found that it shows a more stable and faster convergence with the assistance of $\lossh_{c}$.

\noindent \textbf{End-to-end optimization.}
As mentioned above, our whole pipeline is fully differentiable thus capable of learning both scene coordinates and quality weights, directly from ground truth poses. In the third stage, we train both the scene coordinate network (Fig.~\ref{fig:arch}-A) and quality weight network (Fig.~\ref{fig:arch}-B) using $\lossh$, forcing both of them to learn task-oriented feature representations.

\subsection{Optional RANSAC-like post-processing}
\label{sec:LM-Refine}
The SC-wLS method allows us to directly calculate camera poses with DLT. Although we focus on feed-forward settings, it is still possible to use RANSAC-like post-processing for better results. Specifically, we adopt the Levenberg-Marquardt solver developed by \cite{brachmann2021visual} to post-process DLT poses as an optional step (shortened as \textbf{LM-Refine}). It brings a boost of localization accuracy.%, which even surpasses the results obtained by imposing LM-Refine on RANSAC initial poses. %Meanwhile, DSAC \cite{brachmann2017dsac}\cite{Brachmann2018LearningLI}\cite{brachmann2021visual} also use similar iterative modules called Hypothesis Refinement upon the best RANSAC hypothesis. The performance of this optional refinement algorithm heavily relies on a good initial pose value and suffers in the situations with many outliers.

Specifically, this LM-Refine module is an iterative procedure. It first determines a set of inlier correspondences according to current pose estimate then optimizes poses using the Levenberg–Marquardt algorithm over the inlier set w.r.t. re-projection errors in Eq.~\ref{eq:proj}. This process stops when the number of inlier set converges or until the maximum iteration, which we set to 100 following \cite{brachmann2021visual}.

\subsection{Self-supervised adaptation}
\label{sec:self-sup adaptation}

At test time, we do not have ground truth pose for supervision. However, if the test data is given in the form of image sequences, we could use the photometric loss in co-visible RGB images to supervise our quality weight network Fig.~\ref{fig:arch}-B. We sample two consecutive images $\img^{s}$ and $\img^{t}$ from the test set and synthesize $\tilde{\img^{t}}$ by warping $\img^{s}$ to $\img^{t}$, similar to self-supervised visual odometry methods \cite{zhou2017unsupervised,li2021generalizing}:
\begin{equation}
    p^t \sim \mathbf{K}\tr_t^{-1} \crd^s
    \label{sampling}
\end{equation}
\begin{equation}
    \lossh_{\rm ph}  = \sum_{p} || \img^t(p) - \tilde{\img^{t}}(p)  ||_1 + \lossh_{\rm SSIM}
\end{equation}
where $p$ indexes over pixel coordinates, and $\tilde{\img^{t}}$ is the source view $\img^{s}$ warped to the target frame based on the predicted global scene coordinates $\crd^s$. $\lossh_{\rm SSIM}$ is the structured similarity loss \cite{1284395} imposed on $\img^t(p)$ and $\tilde{\img^{t}}(p)$.

Scene coordinates can be trained by $\lossh_{\rm ph}$ via two gradient paths: (1) the sampling location generation process in Eq.~\ref{sampling}. (2) the absolute pose $\tr_t$ which is calculated by DLT on scene coordinates. However, we observe that fine-tuning scene coordinates with $\lossh_{\rm ph}$ results in divergence. To this end, we detach scene coordinates from computation graphs and only fine-tune the quality weight net (Fig.~\ref{fig:arch}-B) with self supervision. As will be shown later in experiments, this practice greatly improves re-localization performance. A detailed illustration of this adaptation scheme is provided in the supplementary material. 

To clarity, our test-time adaptation experiments exploit all frames in the test set for self-supervision. An ideal online formulation would only use frames before the one of interest. The current offline version can be useful in scenarios like robot vacuums adapts to specific rooms during standby time.
%In conventional self-supervised VO methods \cite{zhou2017unsupervised}, depth and pose estimation are achieved by different networks, thus they do not suffer from this specific problem.

%, including a SVD decomposition operation, which is differential and would back-propagate gradients to scene coordinates. Moreover, benefiting from the pretraining beforehand, the network wouldn't suffer from the eigen vector shifting and can be trained by $\lossh_{un}$.

% A seemingly natural approach to exploiting ED within the network would consist of minimizing the $\ell_2$ loss $\|\be_\theta - \bte\|^2$, where $\bte$ is the ground-truth smallest eigenvector as in~\cite{Ionescu15,Yi18a}. As discussed in Section~\ref{sec:motivation}, however, this requires differentiating the ED result to perform back-propagation and is not optimization-friendly. Furthermore, as argued in~\cite{Hartley00} for the fundamental matrix, the direct use of the $\ell_2$ norm as a distance measure is not optimal, since all the entries do not have equal importance.

% \ok{To avoid the instabilities of eigenvector differentiation and use the algebraic error advocated for in~\cite{Hartley00}, we consider the standard definition of a zero eigen-value $\be_\theta$, i.e., }

% \noindent \textbf{End-to-End Optimization.}

  \begin{table*}[tb]
  	\centering
  	\vspace{0pt}
%   	\small
  	\caption{Median errors on the 7Scenes dataset \cite{shotton2013scene}, with translational and rotational errors measured in $m$ and $^{\circ}$. Our results are evaluated directly using weighted DLT, without the (optional) LM-Refine step (Section \ref{sec:LM-Refine}) and self-supervised adaptation step (Section \ref{sec:self-sup adaptation}). We compare with methods that directly predict poses with neural networks. Ours ($dlt$) and Ours ($dlt$+$e2e$) show results evaluated w/o and w/ the third stage mentioned in Section \ref{sec:training protocol}. Our results are significantly better.}
%   	\vspace{-5pt}
  	 \resizebox{\textwidth}{!}{
  	 
  	\begin{tabular}{lccccccc}
  		\toprule
  		Methods & Chess & Fire & Heads & Office & Pumpkin & Kitchen & Stairs\\
  		\midrule
  		%PoseNet15 \cite{kendall2015posenet} & 0.32/8.1 & 0.47/14.4 & 0.29/12.0 & 0.48/7.7 & 0.47/8.4 & 0.59/8.6 & 0.47/13.8 \\
  		%PoseNet16 \cite{kendall2016modelling} & 0.37/7.2 & 0.43/13.7 & 0.31/12.0 & 0.48/8.0 & 0.61/7.1 & 0.58/7.5 & 0.48/13.1 \\
  		PoseNet17 \cite{kendall2017geometric} & 0.13/4.5 & 0.27/11.3 & 0.17/13.0 & 0.19/5.6 & 0.26/4.8 & 0.23/5.4 & 0.35/12.4 \\
  		LSTM-Pose \cite{walch2017image} & 0.24/5.8 & 0.34/11.9 & 0.21/13.7 & 0.30/8.1 & 0.33/7.0 & 0.37/8.8 & 0.40/13.7 \\
  		BranchNet \cite{wu2017delving} & 0.18/5.2 & 0.34/9.0 & 0.20/14.2 & 0.30/7.1 & 0.27/5.1 & 0.33/7.4 & 0.38/10.3 \\
  		GPoseNet \cite{cai2019hybrid} & 0.20/7.1 & 0.38/12.3 & 0.21/13.8 & 0.28/8.8 & 0.37/6.9 & 0.35/8.2 & 0.37/12.5\\
  		MLFBPPose \cite{wang2019discriminative} & 0.12/5.8 & 0.26/12.0 & 0.14/13.5 & 0.18/8.2 & 0.21/7.1 & 0.22/8.1 & 0.38/10.3 \\
  		AttLoc \cite{wang2020atloc} & 0.10/4.1 & 0.25/11.4 & 0.16/11.8 & 0.17/5.3 & 0.21/4.4 & 0.23/5.4 & 0.26/10.5 \\
  		%AnchorPoint & 0.06/3.9 & 0.16/11.1 & 0.09/11.2 & 0.11/5.4 & 0.14/3.6 & 0.13/5.3 & 0.21/11.9 \\
  		
  		%DSAC \cite{brachmann2017dsac} w/ model & 0.02/1.2 & 0.04/1.5 & 0.03/2.7 & 0.04/1.6 & 0.05/2.0 & 0.05/2.0 & 1.17/33.1  \\
  		%DSAC++ \cite{Brachmann2018LearningLI} w/ model & \underline{0.02}/\underline{0.5} & \underline{0.02}/\underline{0.9} & \underline{0.01}/\underline{0.8} & \underline{0.03}/\underline{0.7} & \underline{0.04}/\underline{1.1} & \underline{0.04}/\underline{1.1} & \underline{0.09}/\underline{2.6} \\
  		\midrule
  		%VidLoc \cite{clark2017vidloc} & 0.18/NA & 0.26/NA & 0.14/NA & 0.26/NA & 0.36/NA & 0.31/NA & 0.26/NA \\
  		MapNet \cite{brahmbhatt2018geometry} & 0.08/3.3 & 0.27/11.7 & 0.18/13.3 & 0.17/5.2 & 0.22/4.0 & 0.23/4.9 & 0.30/12.1\\
  		LsG \cite{xue2019local} & 0.09/3.3 & 0.26/10.9 & 0.17/12.7 & 0.18/5.5 & 0.20/3.7 & 0.23/4.9 & 0.23/11.3 \\ 
  		GL-Net \cite{xue2020learning} & 0.08/2.8 & 0.26/8.9 & 0.17/11.4 & 0.18/5.1 & 0.15/2.8 & 0.25/4.5 & 0.23/8.8\\
  		MS-Transformer \cite{shavit2021learning} & 0.11/4.7 & 0.24/9.6 & 0.14/12.2 & 0.17/5.7 & 0.18/4.4 & 0.17/5.9 & 0.26/8.5 \\
  		\midrule
  		Ours ($dlt$) & 0.029/0.78 & 0.051/\textbf{1.04} & \textbf{0.026}/2.00 & 0.063/0.93 & 0.084/1.28 & 0.099/1.60 & 0.179/3.61 \\
  		Ours ($dlt$+$e2e$) & \textbf{0.029}/\textbf{0.76} & \textbf{0.048}/1.09 & 0.027/\textbf{1.92} & \textbf{0.055}/\textbf{0.86} & \textbf{0.077}/\textbf{1.27} & \textbf{0.091}/\textbf{1.43} & \textbf{0.123}/\textbf{2.80} \\
  		
  		\bottomrule
  	\end{tabular}}
  	\vspace{0pt}
  	\label{e2esota7}
  \end{table*}

\section{Experiments}

\subsection{Experiment setting}
\textbf{Datasets.} We evaluate our SC-wLS framework for camera re-localization from single RGB images, on both indoor and outdoor scenes. Following \cite{brachmann2021visual}, we choose the publicly available indoor 7Scenes dataset \cite{shotton2013scene} and outdoor Cambridge dataset \cite{kendall2015posenet}, which have different scales, appearance and motion patterns. These two datasets have fairly different distributions of scene coordinates. We would show later that, in both cases, SC-wLS successfully predicts interpretable scene coordinate weights for feed-forward camera re-localization.

  \begin{table*}[tb]
  	\centering
  	\small
%   	\vspace{-10pt}
  	\caption{Median errors on the Cambridge dataset \cite{kendall2015posenet}, with translational and rotational errors measured in $m$ and $^{\circ}$. Settings are the same as Table~\ref{e2esota7}. Our results are significantly better than other methods. Note that GL-Net uses a sequence as input. 
  	}
  	\resizebox{100mm}{!}{
  	\begin{tabular}{lccccc}
  		%\begin{tabular}{cm|cm|cm|cm|cm|cm|cm|cm|cm|cm|cm|cm|cm}
  		\toprule
  		Methods & Greatcourt & King's College & Shop Facade & Old Hospital & Church\\
  		\midrule
  		% 		InLoc \cite{taira2018inloc} & 1.20/0.6 & 0.46/0.8 & 0.11/0.5 & 0.48/1.0 & 0.18/0.6 \\
  		
  		% 		BTBRF \cite{meng2017backtracking} & N/A & 0.39/0.4 & 0.15/0.3 & 0.30/0.4 & 0.20/0.4 \\
  		% 		SANet \cite{yang2019sanet} & 3.28/2.0 & 0.32/0.5 & 0.10/0.5 & 0.32/0.5 & 0.16/0.6 \\
  		ADPoseNet \cite{huang2019prior} & N/A & 1.30/1.7 & 1.22/6.7 & N/A & 2.28/4.8 \\
  		%PoseNet15 \cite{kendall2015posenet} & N/A & 1.66/4.9 & 1.41/7.2 & 2.62/4.9 & 2.45/8.0 \\
  		%PoseNet16 \cite{kendall2016modelling} & N/A & 1.74/4.1 & 1.25/7.5 & 2.57/5.1 & 1.74/4.1 \\
  		PoseNet17 \cite{kendall2017geometric} & 7.00/3.7 & 0.99/1.1 & 1.05/4.0 & 2.17/2.9 & 1.49/3.4 \\
  		GPoseNet \cite{cai2019hybrid} & N/A & 1.61/2.3 & 1.14/5.7 & 2.62/3.9 & 2.93/6.5 \\
  		MLFBPPose \cite{wang2019discriminative} & N/A & 0.76/1.7 & 0.75/5.1 & 1.99/2.9 & 1.29/5.0 \\
  		
  		LSTM-Pose \cite{walch2017image} & N/A & 0.99/3.7 & 1.18/7.4 & 1.51/4.3 & 1.52/6.7 \\
  		SVS-Pose \cite{naseer2017deep} & N/A & 1.06/2.8 & 0.63/5.7 & 1.50/4.0 & 2.11/8.1 \\
  		%AnchorNet \cite{saha2018improved} & N/A & 0.57/0.9 & 0.52/2.3 & 1.21/2.6 & 1.04/2.7\\
  		
  		\midrule
  		MapNet \cite{brahmbhatt2018geometry} & N/A & 1.07/1.9 & 1.49/4.2 & 1.94/3.9 & 2.00/4.5 \\
  		GL-Net \cite{xue2020learning} & 6.67/2.8 & 0.59/0.7 & 0.50/2.9 & 1.88/2.8 & 1.90/3.3\\
  		MS-Transformer \cite{shavit2021learning} & N/A & 0.83/1.5 & 0.86/3.1 & 1.81/2.4 & 1.62/4.0 \\
  		\midrule
  		Ours ($dlt$) & 1.81/1.2 & 0.22/0.9 & 0.15/1.1 & 0.46/1.9 & 0.50/1.5 \\
  		Ours ($dlt$+$e2e$) & \textbf{1.64}/\textbf{0.9} & \textbf{0.14}/\textbf{0.6} & \textbf{0.11}/\textbf{0.7} & \textbf{0.42}/\textbf{1.7} & \textbf{0.39}/\textbf{1.3} \\ 
  		\bottomrule
  	\end{tabular}
  	}
  	\label{e2esota_cam}
  \end{table*}

\noindent \textbf{Implementation.} The input images are proportionally resized so that the heights are 480 pixels. We randomly zoom and rotate images for data augmentation following \cite{brachmann2021visual}. As mentioned in Sec. \ref{sec:loss func}, $\alpha$ and $\beta$ in Eq.~\ref{eq:ell_comb} are sensitive to specific scenes. We set them to 5 and 1e-4 for indoor 7Scenes while 5 and 1e-6 for outdoor Cambridge, respectively. 
% (A slight tuning according to specific scenes will enhance the performance in some degrees. Generally, decreasing the value of these parameters slightly when training will lead to an enhancement of pose estimation accuracy. Supplementary?) 
The balancing hyper-parameter $\gamma$ is set to 5 empirically. We train our model on one nVIDIA GeForce RTX 3090 GPU. The ADAM optimizer with initial learning rate 1e-4 is utilized in the first two training stages and the learning rate is set to 1e-5 in the end-to-end training stage. The batch size is set to 1 as \cite{brachmann2021visual}. The scene coordinate regression network architecture for 7Scenes is adopted from \cite{brachmann2021visual}. As for Cambridge, we add residual connections to the early layers of this network. Architecture details can be found in the supplementary material.

\subsection{Feed-forward Re-localization}
\label{sec:direct re-localization}

An intriguing property of the proposed SC-wLS method is that we can re-localize a camera directly using weighted least squares. This inference scheme seamlessly blends into the forward pass of a neural network. So we firstly compare with other APR re-localization methods that predict camera poses directly with a neural network during inference. Quantitative results on 7Scenes and Cambridge are summarized in Table~\ref{e2esota7} and Table~\ref{e2esota_cam}, respectively. 

\begin{table*}[tb]

	\centering
% 	\small
	\caption{Results on the 7Scenes dataset \cite{shotton2013scene} and the Cambrdige dataset \cite{kendall2015posenet}, with translational and rotational errors measured in $m$ and $^{\circ}$. Here $ref$ means LM-Refine. Note DSAC* \cite{brachmann2021visual} w/o model and NG-RANSAC \cite{Brachmann2019NeuralGuidedRL} also use the LM pose refinement process. Note that with \emph{ref} used, our method loses the feed-forward nature. NG-RANSAC w/o model$\dag$ is retrained using DSAC* RGB initialization.}
	%\begin{tabular}{p{0.47cm}|p{0.47cm}|p{0.47cm}|cp{0.47cm}|cp{0.47cm}|cp{0.47cm}|cp{0.47cm}|cp{0.47cm}|cp{0.47cm}|cp{0.47cm}|cp{0.47cm}|cp{0.47cm}|cp{0.47cm}|cp{0.47cm}|cp{0.47cm}|cp{0.47cm}|cp{0.47cm}|cp{0.47cm}|cp{0.47cm}}
	 \resizebox{\textwidth}{!}{
	 
	\begin{tabular}{lccccccc}
		%\begin{tabular}{cm|cm|cm|cm|cm|cm|cm|cm|cm|cm|cm|cm|cm}
		\toprule
		7Scenes & Chess & Fire & Heads & Office & Pumpkin & Kitchen & Stairs\\
		\midrule
		%DSAC \cite{brachmann2017dsac} w/ model & 0.02/1.2 & 0.04/1.5 & 0.03/2.7 & 0.04/1.6 & 0.05/2.0 & 0.05/2.0 & 1.17/33.1  \\
		%DSAC++ \cite{Brachmann2018LearningLI} w/ model & \underline{0.02}/\underline{0.5} & \underline{0.02}/\underline{0.9} & \underline{0.01}/\underline{0.8} & \underline{0.03}/\underline{0.7} & \underline{0.04}/\underline{1.1} & \underline{0.04}/\underline{1.1} & \underline{0.09}/\underline{2.6} \\
		%KFNet \cite{zhou2020kfnet} & 0.018/0.65 & 0.023/0.9 & 0.014/0.82 & 0.025/0.69 & 0.037/1.02 & 0.038/1.16 & 0.033/0.94 \\
		%VS-Net \cite{huang2021vs} & 0.015/0.5 & 0.019/0.8 & 0.012/0.7 & 0.021/0.6 & 0.037/1.0 & 0.036/1.1 & 0.028/0.8 \\
		%PixLoc(RPR) & 0.03/0.91 & 0.02/0.87 & 0.01/0.77 & 0.03/0.94 & 0.05/1.14 & 0.04/1.44 & 0.06/1.38 \\
		%DSAC++ \cite{Brachmann2018LearningLI} w/o model & 0.02/0.7 & 0.03/1.1 & 0.12/6.7 & 0.03/\textbf{0.8} & 0.05/1.1 & 0.05/1.3 & 0.29/5.1 \\
		DSAC* \cite{brachmann2021visual} w/o model & 0.019/1.11 & \textbf{0.019}/1.24 & \textbf{0.011}/1.82 & \textbf{0.026}/1.18 & \textbf{0.042}/1.42 & \textbf{0.030}/1.70 & \textbf{0.041}/\textbf{1.42} \\
		\midrule
		% 		Baseline & 0.029/0.78 & 0.051/1.04 & 0.026/2 & 0.074/0.99 & 0.097/1.37 & 0.108/1.81 & 0.179/3.61 \\
		% 		Baseline+e2e & 0.029/0.76 & 0.05/1.05 & 0.027/1.92 & 0.056/0.9 & 0.08/1.25 & 0.102/1.62 & 0.12/2.57 \\
		% 		Baseline+refine & 0.018/0.63 & 0.026/0.84 & 0.015/0.87 & 0.038/0.85 & 0.045/1.04 & 0.052/1.19 & 0.0672/1.93 \\
		% 		Baseline+e2e+refine & 0.019/0.62 & 0.026/0.9 & 0.014/0.9 & 0.037/0.82 & 0.052/1.14 & 
		% 		\midrule
		%Ours($dlt$+$e2e$) & 0.029/0.76 & 0.05/1.05 & 0.027/1.92 & 0.056/0.9 & 0.08/1.25 & 0.102/1.62 & 0.12/2.57 \\
		Ours ($dlt$+$ref$) & \textbf{0.018}/0.63 & 0.026/\textbf{0.84} & 0.015/\textbf{0.87} & 0.038/0.85 & 0.045/\textbf{1.05} & 0.051/1.17 & 0.067/1.93 \\
		Ours ($dlt$+$e2e$+$ref$) & 0.019/\textbf{0.62} & 0.025/0.88 & 0.014/0.90 & 0.035/\textbf{0.78} & 0.051/1.07 & 0.054/\textbf{1.15} & 0.058/1.57 \\
		\midrule
		Cambridge & & Greatcourt & King's College & Shop Facade & Old Hospital & Church\\
		\midrule
		% 		PoseNet15 \cite{kendall2015posenet} & N/A & 1.66/4.9 & 1.41/7.2 & 2.62/4.9 & 2.45/8.0 \\
		% 		InLoc \cite{taira2018inloc} & 1.20/0.6 & 0.46/0.8 & 0.11/0.5 & 0.48/1.0 & 0.18/0.6 \\
		%Active Search \cite{sattler2016efficient} & N/A & 0.42/0.6 & 0.12/0.4 & 0.44/1.0 & 0.19/0.5 \\
		% 		BTBRF \cite{meng2017backtracking} & N/A & 0.39/0.4 & 0.15/0.3 & 0.30/0.4 & 0.20/0.4 \\
		% 		SANet \cite{yang2019sanet} & 3.28/2.0 & 0.32/0.5 & 0.10/0.5 & 0.32/0.5 & 0.16/0.6 \\
		%DSAC \cite{brachmann2017dsac} & 2.80/1.5 & 0.30/0.5 & 0.09/0.4 & 0.33/0.6 & 0.55/1.6\\
		%DSAC++ \cite{Brachmann2018LearningLI} w/ model & 0.40/0.2 & 0.18/0.3 & 0.06/0.3 & 0.20/0.3 & 0.13/0.4\\
		%DSAC* \cite{brachmann2021visual} w/ model & 0.49/0.3 & 0.15/0.3 & 0.05/0.3 & 0.21/0.4 & 0.13/0.4		\\
		%KFNet \cite{zhou2020kfnet} & 0.42/0.2 & 0.16/0.3 & \underline{0.05}/\underline{0.3} & \underline{0.18},\underline{0.3} & 0.12/0.4 \\
		
		%\midrule
		%VS-Net \cite{huang2021vs} & 0.22/0.1 & 0.16/0.2 & 0.06/0.3 & 0.16/0.3 & 0.08/0.3 \\
		%\midrule
		%PixLoc(RPR) & 0.30/0.14 & 0.14/0.24 & 0.05/0.23 & 0.16/0.32 & 0.10/0.34 \\
		%\midrule
		%DSAC++ \cite{Brachmann2018LearningLI} w/o model & 0.66/0.4 & 0.23/0.4 & 0.09/0.4 & 0.24/0.5 & 0.2/0.7 \\
		DSAC* \cite{brachmann2021visual} w/o model & &0.34/0.2 & 0.18/0.3 & 0.05/1.3 & 0.21/0.4 & 0.15/0.6 \\
		NG-RANSAC \cite{Brachmann2019NeuralGuidedRL} & & 0.35/0.2 & 0.13/0.2 & 0.06/0.3 & 0.22/0.4 & 0.10/0.3 \\
		NG-RANSAC w/o model\dag & & 0.31/0.2 & 0.15/0.3 & 0.05/0.3 & 0.20/0.4 & 0.12/0.4 \\
		\midrule
		%Ours($dlt$+$e2e$) & 14.01/14.0 & 0.24/10 & 0.21/1.6 & 0.72/3.0 & 0.53/2.0 \\ 
		Ours ($dlt$+$ref$) & & 0.32/0.2 & 0.09/0.3 & 0.04/0.3 & 0.12/0.4 & 0.12/0.4 \\
		Ours ($dlt$+$e2e$+$ref$) & & \textbf{0.29}/\textbf{0.2} & \textbf{0.08}/\textbf{0.2} & \textbf{0.04}/\textbf{0.3} & \textbf{0.11}/\textbf{0.4} & \textbf{0.09}/\textbf{0.3} \\
		%ours w/ 50\% RANSAC & 0.02/1.71 & 0.035/1.13 & 0.018/1.02 & 0.04/0.94 & & & 0.06/1.54\\
		% 		ours w/ LM-refine & 0.019/0.64 & 0.027/0.90 & 0.0149/0.88 & & 0.046/1.07 & & 0.07/2.04\\
		%\midrule
		%Ours($dlt$+$e2e$+$self$) & 0.025/0.71 & 0.033,0.98 & 0.026/1.84 & 0.041/0.81 & 0.061/1.21 & 0.062/0.26 & 0.068/1.24 \\
		
		\bottomrule
	\end{tabular}}
	%\setlength{\abovecaptionskip}{-10cm}
	%\setlength{\belowcaptionskip}{-10cm}
% 	\vspace{7pt}
	\label{sota7}
	\vspace{0pt}
\end{table*}

As for former arts, we distinguish between two cases: single frame based and sequence based. They are separated by a line in Table~\ref{e2esota7} and Table~\ref{e2esota_cam}. As for our results, we report under two settings: Ours ($dlt$) and Ours ($dlt$+$e2e$). Ours ($dlt$+$e2e$) means all three training stages are finished (see Section \ref{sec:training protocol}), while Ours ($dlt$) means only the first two stages are used.

On 7Scenes, it is clear that our SC-wLS method out-performs APR camera re-localization methods by significant margins. Note we only take a single image as input, and still achieve much lower errors than sequence-based methods like GLNet \cite{xue2020learning}. On Cambrige, SC-wLS also reports significantly better results including the Greatcourt scene. Note that this scene has extreme illumination conditions and most former solutions do not even work.

%   \setlength{\tabcolsep}{1.4pt}
%   \vspace{-20pt}
\begin{table*}[tb]
	\centering
% 	\small
    \vspace{0pt}
	\caption{Recalls (\%) on the 7Scenes dataset \cite{shotton2013scene} and the Cambrdige dataset \cite{kendall2015posenet}.}
	%\begin{tabular}{p{0.47cm}|p{0.47cm}|p{0.47cm}|cp{0.47cm}|cp{0.47cm}|cp{0.47cm}|cp{0.47cm}|cp{0.47cm}|cp{0.47cm}|cp{0.47cm}|cp{0.47cm}|cp{0.47cm}|cp{0.47cm}|cp{0.47cm}|cp{0.47cm}|cp{0.47cm}|cp{0.47cm}|cp{0.47cm}|cp{0.47cm}}
	 \resizebox{90mm}{!}{
	 
	\begin{tabular}{lcccccccc}
		%\begin{tabular}{cm|cm|cm|cm|cm|cm|cm|cm|cm|cm|cm|cm|cm}
		\toprule
		7Scenes & Chess & Fire & Heads & Office & Pumpkin & Kitchen & Stairs & Avg\\
		\midrule
		%DSAC \cite{brachmann2017dsac} w/ model & 0.02/1.2 & 0.04/1.5 & 0.03/2.7 & 0.04/1.6 & 0.05/2.0 & 0.05/2.0 & 1.17/33.1  \\
		%DSAC++ \cite{Brachmann2018LearningLI} w/ model & \underline{0.02}/\underline{0.5} & \underline{0.02}/\underline{0.9} & \underline{0.01}/\underline{0.8} & \underline{0.03}/\underline{0.7} & \underline{0.04}/\underline{1.1} & \underline{0.04}/\underline{1.1} & \underline{0.09}/\underline{2.6} \\
		
		DSAC* \cite{brachmann2021visual} w/o model & 96.7 & 92.9 & \textbf{98.2} & 87.1 & 60.7 & 65.3 & 64.1 & \textbf{80.7} \\
		Active Search \cite{sattler2016efficient} & 86.4 & 86.3 & 95.7 & 65.6 & 34.1 & 45.1 & \textbf{67.8} & 68.7 \\
		\midrule
		
		Ours ($dlt$+$e2e$+$ref$) & 93.7 & 82.2 & 65.7 & 73.5 & 49.6 & 45.3 & 43.0 & 64.7 \\
		Ours ($dlt$+$e2e$+$dsac*$) & \textbf{96.8} & \textbf{97.4} & 94.3 & \textbf{88.6} & \textbf{62.4} & \textbf{65.5} & 55.3 & 80.0 \\
		\midrule
		Cambridge & & Great & King's & Shop & Old & Church & Avg\\
		& & Court & College & Facade &  Hospital &  & \\
		\midrule
		% 		PoseNet15 \cite{kendall2015posenet} & N/A & 1.66/4.9 & 1.41/7.2 & 2.62/4.9 & 2.45/8.0 \\
		% 		InLoc \cite{taira2018inloc} & 1.20/0.6 & 0.46/0.8 & 0.11/0.5 & 0.48/1.0 & 0.18/0.6 \\
		%Active Search \cite{sattler2016efficient} & N/A & 0.42/0.6 & 0.12/0.4 & 0.44/1.0 & 0.19/0.5 \\
		% 		BTBRF \cite{meng2017backtracking} & N/A & 0.39/0.4 & 0.15/0.3 & 0.30/0.4 & 0.20/0.4 \\
		% 		SANet \cite{yang2019sanet} & 3.28/2.0 & 0.32/0.5 & 0.10/0.5 & 0.32/0.5 & 0.16/0.6 \\
		%DSAC \cite{brachmann2017dsac} & 2.80/1.5 & 0.30/0.5 & 0.09/0.4 & 0.33/0.6 & 0.55/1.6\\
		%DSAC++ \cite{Brachmann2018LearningLI} w/ model & 0.40/0.2 & 0.18/0.3 & 0.06/0.3 & 0.20/0.3 & 0.13/0.4\\
		%DSAC* \cite{brachmann2021visual} w/ model & 0.49/0.3 & 0.15/0.3 & 0.05/0.3 & 0.21/0.4 & 0.13/0.4		\\
		%KFNet \cite{zhou2020kfnet} & 0.42/0.2 & 0.16/0.3 & \underline{0.05}/\underline{0.3} & \underline{0.18},\underline{0.3} & 0.12/0.4 \\
		
		%\midrule
		%VS-Net \cite{huang2021vs} & 0.22/0.1 & 0.16/0.2 & 0.06/0.3 & 0.16/0.3 & 0.08/0.3 \\
		%\midrule
		%PixLoc(RPR) & 0.30/0.14 & 0.14/0.24 & 0.05/0.23 & 0.16/0.32 & 0.10/0.34 \\
		%\midrule
		%DSAC++ \cite{Brachmann2018LearningLI} w/o model & 0.66/0.4 & 0.23/0.4 & 0.09/0.4 & 0.24/0.5 & 0.2/0.7 \\
		DSAC* \cite{brachmann2021visual} w/o model & & 62.9 & 80.8 & 86.4 & 55.5 & 88.9 & 74.9 \\
		NG-RANSAC w/o model\dag & & 67.6 & 81.9 & 88.3 & 56.0 & 93.8 & 77.5 \\
		\midrule
		%Ours($dlt$+$e2e$) & 14.01/14.0 & 0.24/10 & 0.21/1.6 & 0.72/3.0 & 0.53/2.0 \\ 
		Ours ($dlt$+$e2e$+$ref$) & & 73.3 & \textbf{97.3} & \textbf{90.3} & \textbf{75.3} & 82.2 & \textbf{83.7} \\
		Ours ($dlt$+$e2e$+$dsac*$) & & \textbf{74.3} & 87.2 & 87.4 & 53.8 & \textbf{99.1} & 80.4 \\
		%ours w/ 50\% RANSAC & 0.02/1.71 & 0.035/1.13 & 0.018/1.02 & 0.04/0.94 & & & 0.06/1.54\\
		% 		ours w/ LM-refine & 0.019/0.64 & 0.027/0.90 & 0.0149/0.88 & & 0.046/1.07 & & 0.07/2.04\\
		%\midrule
		%Ours($dlt$+$e2e$+$self$) & 0.025/0.71 & 0.033,0.98 & 0.026/1.84 & 0.041/0.81 & 0.061/1.21 & 0.062/0.26 & 0.068/1.24 \\
		
		\bottomrule
	\end{tabular}}
	\vspace{0pt}
	\label{recall}
\end{table*}

\textbf{Why SC-wLS performs much better than former APR methods that directly predict poses without RANSAC?} Because they are black boxes, failing to model explicit geometric relationships. These black-box methods suffer from hard memorization and dataset bias. By contrast, our formulation is based upon 2D-3D correspondences, while still allowing feed-forward inference. 

Lastly, 
% our \textit{dlt+e2e}
Ours ($dlt$+$e2e$) 
shows lower errors than Ours ($dlt$) in most scenes, validating the effectiveness of jointly fine-tuning Fig.~\ref{fig:arch}-A and Fig.~\ref{fig:arch}-B.

\subsection{Optional LM-Refine}

The most interesting part about SC-wLS is the strong re-localization ability without using iterative pose estimation methods like RANSAC or LM-Refine. However, although optional, incorporating such methods does lead to lower errors. We report quantitative comparisons in Table~\ref{sota7}. We compare with the best published method DSAC* in the 'w/o model' setting for two reasons: 1) Our training protocol does not involve any usage of 3D models; 2) DSAC* also uses LM-Refine as a post-processing step. Note LM-Refine needs pose initialization, so we use DLT results as initial poses. Similar to above experiments, Ours ($dlt$+$e2e$+$ref$) means all three training stages are used. For Cambridge, we also compare to NG-RANSAC \cite{Brachmann2019NeuralGuidedRL}, which predicts weights from RGB images.

On the Cambridge dataset, SC-wLS outperforms state-of-the-art methods. On the Old Hospital scene, we reduce the translational error from 0.21m to 0.11m, which is a 47.6\% improvement. Meanwhile, Ours ($dlt$+$e2e$+$ref$) consistently achieves lower errors than Ours ($dlt$+$ref$), showing the benefits of end-to-end joint learning. On the 7Scenes dataset, our results under-perform DSAC* and the third training stage does not bring clear margins.

\begin{table*}[tb]
	\centering
% 	\small
% \scriptsize
\vspace{0pt}
	\caption{Results on the 7Scenes dataset \cite{shotton2013scene} and the Cambrdige dataset \cite{kendall2015posenet}, with translational and rotational errors measured in $m$ and $^{\circ}$. Here $self$ means self-supervised adaptation. Note these results are evaluated without LM-Refine.}
	%\begin{tabular}{p{0.47cm}|p{0.47cm}|p{0.47cm}|cp{0.47cm}|cp{0.47cm}|cp{0.47cm}|cp{0.47cm}|cp{0.47cm}|cp{0.47cm}|cp{0.47cm}|cp{0.47cm}|cp{0.47cm}|cp{0.47cm}|cp{0.47cm}|cp{0.47cm}|cp{0.47cm}|cp{0.47cm}|cp{0.47cm}|cp{0.47cm}}
	 \resizebox{\textwidth}{!}{
	 
	\begin{tabular}{lccccccc}
		%\begin{tabular}{cm|cm|cm|cm|cm|cm|cm|cm|cm|cm|cm|cm|cm}
		\toprule
		7Scenes & Chess & Fire & Heads & Office & Pumpkin & Kitchen & Stairs\\
		\midrule
		
		Ours ($dlt$+$e2e$) & 0.029/0.76 & 0.048/1.09 & 0.027/1.92 & 0.055/0.86 & 0.077/1.27 & 0.091/1.43 & 0.123/2.80 \\
		
% 		Ours ($dlt$+$e2e$+$self$)(150k) & \textbf{0.025}/\textbf{0.71} & \textbf{0.033}/\textbf{0.98} & \textbf{0.026}/\textbf{1.84} & \textbf{0.041}/\textbf{0.81} & \textbf{0.061}/\textbf{1.21} & \textbf{0.062}/\textbf{1.26} & \textbf{0.068}/\textbf{1.24} \\
        Ours ($dlt$+$e2e$+$self$) & \textbf{0.021}/\textbf{0.64} & \textbf{0.023}/\textbf{0.80} & \textbf{0.013}/\textbf{0.79} & \textbf{0.037}/\textbf{0.76} & \textbf{0.048}/\textbf{1.06} & \textbf{0.051}/\textbf{1.08} & \textbf{0.055}/\textbf{1.48} \\
		\midrule
		Cambridge & & Greatcourt & King's College & Shop Facade & Old Hospital & Church\\
		\midrule
		Ours ($dlt$+$e2e$) & & 1.64/0.9 & 0.14/0.6 & 0.11/0.7 & 0.42/1.7 & 0.39/1.3 \\ 
		Ours ($dlt$+$e2e$+$self$) & & \textbf{0.94}/\textbf{0.5} & \textbf{0.11}/\textbf{0.3} & \textbf{0.05}/\textbf{0.4} & \textbf{0.18}/\textbf{0.7} & \textbf{0.17}/\textbf{0.8} \\
		\bottomrule
	\end{tabular}}
 	\vspace{0pt}
	\label{self_all}
\end{table*}

We also report average recalls with pose error below 5cm,
5deg (7Scenes) and translation error below 0.5\% of the
scene size (Cambridge) in Table~\ref{recall}. It is demonstrated that the recall value of Ours ($dlt$+$e2e$+$ref$) under-performs SOTA on 7Scenes, which is consistent with the median error results. We also evaluate DSAC*'s exact post-processing, denoted as Our ($dlt$+$e2e$+$dsac*$). The difference is that for Ours  ($dlt$+$e2e$+$ref$) we use DLT results for LM-Refine initialization and for Ours ($dlt$+$e2e$+$dsac*$) we use RANSAC for LM-Refine initialization. It is shown that using DSAC*'s exact post-processing can compensate for the performance gap on 7Scenes.

\subsection{Self-supervised Adaptation}

We show the effectiveness of self-supervised weight network adaptation during test time, which is a potentially useful new feature of SC-wLS (Section \ref{sec:self-sup adaptation}). Results are summarized in Table~\ref{self_all}, which is evaluated under the weighted DLT setting (same as Section \ref{sec:direct re-localization}). Obviously, for all sequences in 7Scenes and Cambridge, Ours ($dlt$+$e2e$+$self$) outperforms Ours ($dlt$+$e2e$) by clear margins. On \emph{Stairs} and \emph{Old Hospital}, translational and rotational error reductions are both over 50\%. In these experiments, self-supervised adaptation runs for about 600k iterations. Usually, this adaptation process converges to a reasonably good state, within only 150k iterations. More detailed experiments are in the supplementary.

% \begin{table}
% 	\centering
% 	\small
% 	\caption{Average end-to-end training time per iteration (s/iter).}
% % \footnotesize
% 	%\begin{tabular}{p{0.47cm}|p{0.47cm}|p{0.47cm}|cp{0.47cm}|cp{0.47cm}|cp{0.47cm}|cp{0.47cm}|cp{0.47cm}|cp{0.47cm}|cp{0.47cm}|cp{0.47cm}|cp{0.47cm}|cp{0.47cm}|cp{0.47cm}|cp{0.47cm}|cp{0.47cm}|cp{0.47cm}|cp{0.47cm}|cp{0.47cm}}
% % 	 \resizebox{\textwidth}{1mm}
%     % \setlength{\tabcolsep}{1.4mm}
%     \resizebox{50mm}{!}{
% 	\begin{tabular}{c|c|c|c}
% 		%\begin{tabular}{cm|cm|cm|cm|cm|cm|cm|cm|cm|cm|cm|cm|cm}
% 		\toprule
% 		Setting & Ours & DSAC* & NG-RANSAC   \\
% 		\midrule
% 		1-GPU & \textbf{0.09} & 0.20 & 0.31   \\
%         5-GPU & \textbf{0.09} & 0.85 & 1.05  \\
% 		\bottomrule
% 	\end{tabular}
% 	}
% 	\label{training}
% \end{table}

\begin{figure*}[tb]
	\begin{center}
\includegraphics[width=0.98\linewidth]{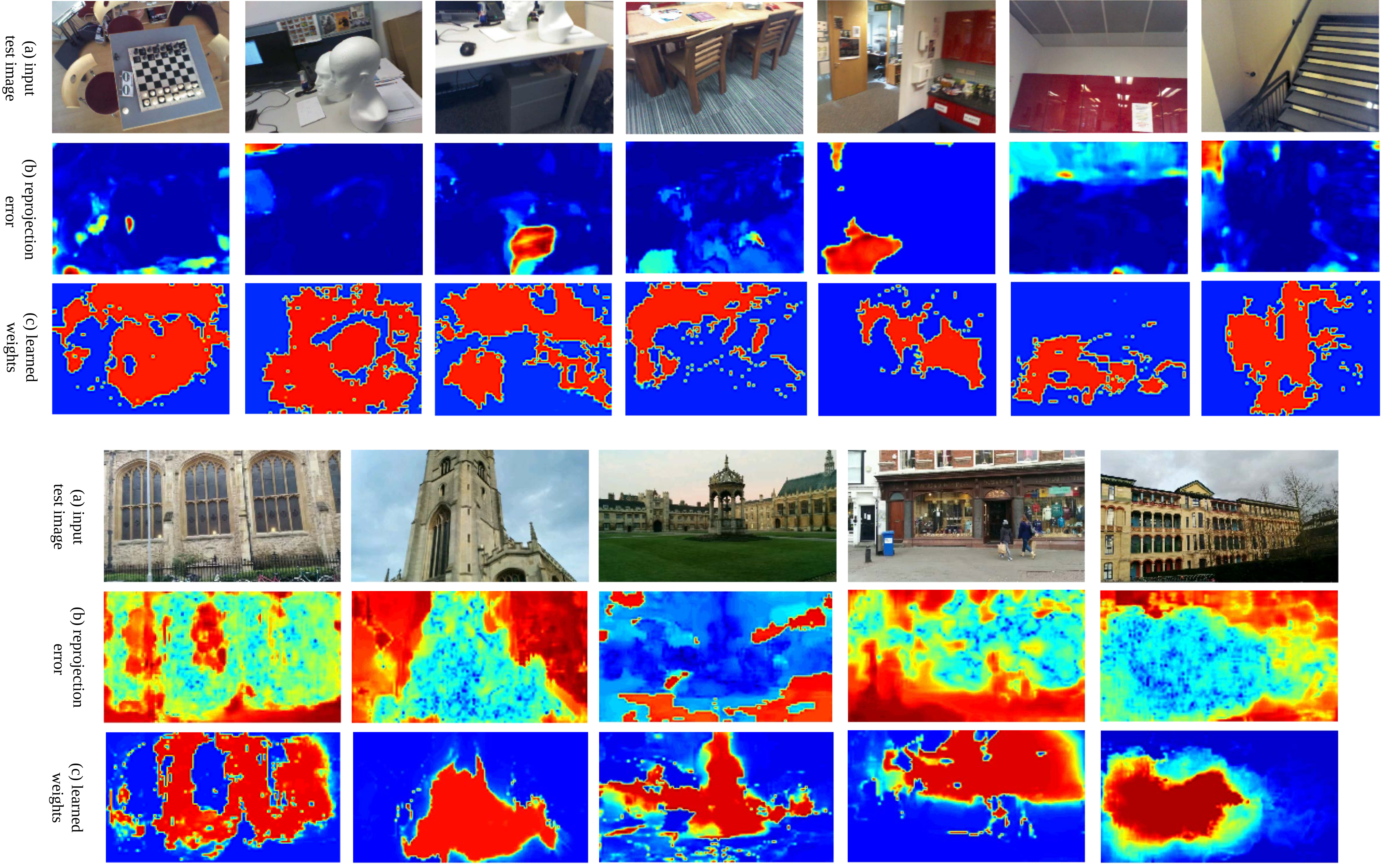}
	\end{center}
	\vspace{-10pt}
	\caption{Visualizations on 7Scenes and Cambridge test sets, demonstrating that our network learns interpretable scene coordinate weights consistent with re-projection errors, by solely considering the intrinsic structure of input 2D-3D correspondences. A higher color temperature represents a higher value.}
	\label{fig:weight}
\end{figure*}

\subsection{Visualization}

Firstly, we demonstrate more learnt weights on the test sets of 7Scenes and Cambridge, in Fig.~\ref{fig:weight}. The heatmaps for reprojection error and learnt weight are highly correlated. Pixels with low quality weights usually have high reprojection errors and occur in sky or uniformly textured regions.

\begin{figure}[tb]
\vspace{0pt}
	\begin{center}
		\includegraphics[scale=0.33]{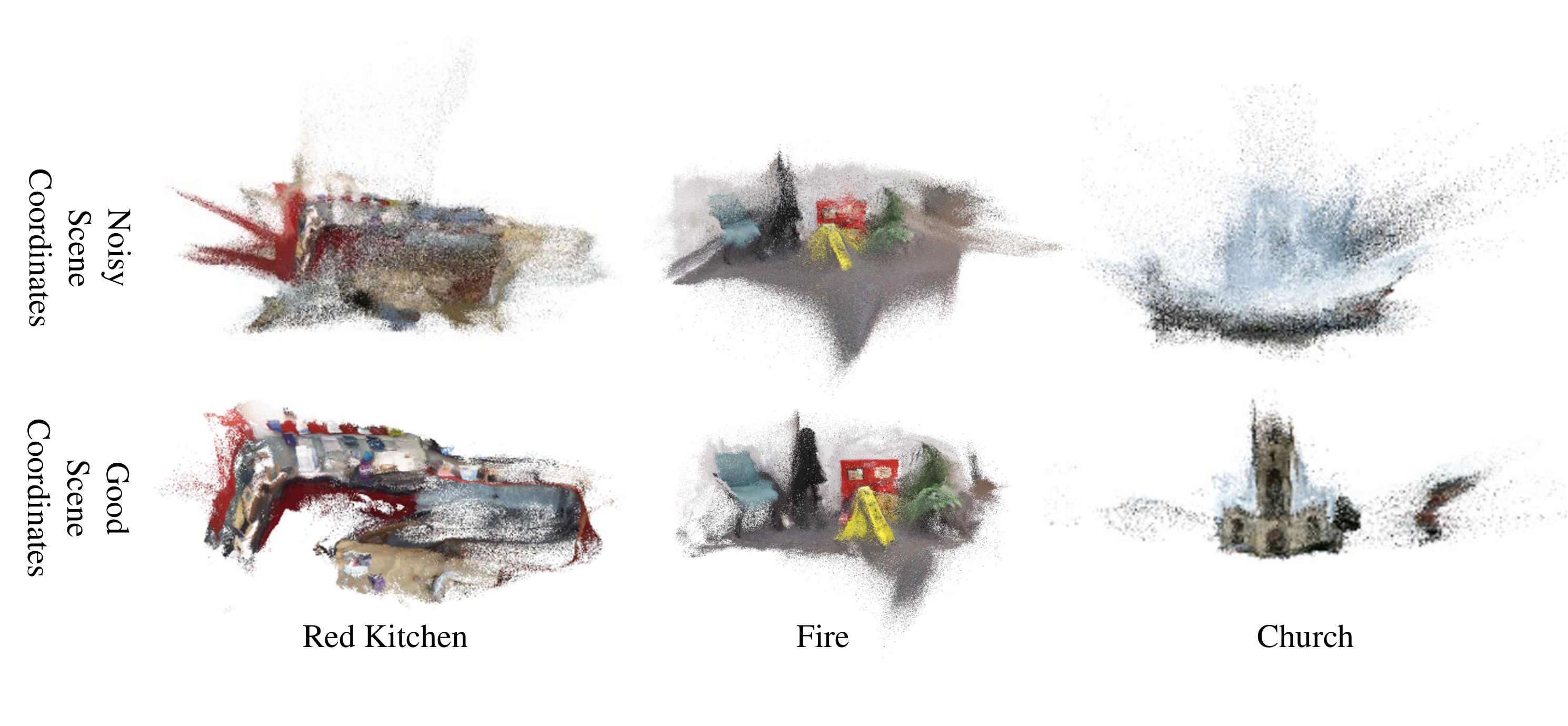}
	\end{center}
	\vspace{-20pt}
	\caption{Map visualization on 7Scenes and Cambridge. More comparisons are provided in the supplementary material.}
	\label{fig:map}
\end{figure}

Secondly, We show learnt 3D maps with and without quality filtering, in Fig.~\ref{fig:map}. Since scene coordinates are predicted in the world frame, we directly show the point clouds generated by aggregating scene coordinate predictions on test frames. It is shown that only predicting scene coordinates results in noisy point clouds, especially in outdoor scenes where scene coordinate predictions on sky regions are only meaningful in term of their 2D projections. We show good scene coordinates by filtering out samples with a quality weight lower than 0.9.

\subsection{Training and Inference Efficiency}

As shown in Table~\ref{timeall}, our method, as a feed-forward (\emph{fw}) one, is faster than the well-engineered iterative (\emph{iter}) method DSAC* and the transformer-based method MS-Transformer \cite{shavit2021learning}. We could make a tradeoff using OANet w/o attention for even faster speed.
Thus it's reasonable to compare our feed-forward method with APR methods. We also evaluate end-to-end training efficiency. When training 5 network instances for 5 different scenes on 5 GPUs, the average time of ours stays unchanged, while that of DSAC* increases due to limited CPU computation/scheduling capacity. We believe large-scale training of many re-localization network instances on the cloud is an industry demand.

\begin{table}[tb]
	\centering
% 	\small
\vspace{0pt}
	\caption{Average end-to-end training and inference time comparisons.}
    % \footnotesize
	%\begin{tabular}{p{0.47cm}|p{0.47cm}|p{0.47cm}|cp{0.47cm}|cp{0.47cm}|cp{0.47cm}|cp{0.47cm}|cp{0.47cm}|cp{0.47cm}|cp{0.47cm}|cp{0.47cm}|cp{0.47cm}|cp{0.47cm}|cp{0.47cm}|cp{0.47cm}|cp{0.47cm}|cp{0.47cm}|cp{0.47cm}|cp{0.47cm}}
% 	 \resizebox{\textwidth}{1mm}
    % \setlength{\tabcolsep}{0.4mm}
    \vspace{0pt}
    \resizebox{110mm}{!}{
	\begin{tabular}{c|c|c|c|c|c|c|c c|c|c|c}
		%\begin{tabular}{cm|cm|cm|cm|cm|cm|cm|cm|cm|cm|cm|cm|cm}
% 		\toprule
        \hline
        \multicolumn{7}{c|}{Inference time (ms/frame)} & & \multicolumn{4}{c}{End-to-end training time (s/frame)} \\
        \hline
		Method & AtLoc & MapNet & Ours & Ours & MS-T & DSAC* & & Setting & Ours & DSAC* & NG-RANSAC   \\
		& \cite{wang2020atloc} & \cite{brahmbhatt2018geometry} & (OANet) &  & \cite{shavit2021learning} & \cite{brachmann2021visual} & &  &  & \cite{brachmann2021visual} & \cite{Brachmann2019NeuralGuidedRL}   \\
% 		\midrule
        \hline
		Type & \emph{fw} & \emph{fw} & \emph{fw} & \emph{fw} & \emph{fw} & \emph{iter} & & 1-GPU & \textbf{0.09} & 0.20 & 0.31   \\
% 		\midrule
        \hline
        Time & \textbf{6} & 9 & 13 & 19 & 30 & 32 & & 5-GPU & \textbf{0.09} & 0.85 & 1.05    \\
        \hline
% 		\bottomrule
	\end{tabular}
	}
	\vspace{0pt}
	\label{timeall}
\end{table}

\section{Conclusions}
In this study, we propose a new camera re-localization solution named SC-wLS, which combines the advantages of feed-forward formulations and scene coordinate based methods. It exploits the correspondences between pixel coordinates and learnt 3D scene coordinates, while still allows direct camera re-localization through a single forward pass. This is achieved by a correspondence weight network that finds high-quality scene coordinates, supervised by poses only. 
Meanwhile, SC-wLS also allows self-supervised test-time adaptation.
Extensive evaluations on public benchmarks 7Scenes and Cambridge demonstrate the effectiveness and interpretability of our method. In the feed-forward setting, SC-wLS results are significantly better than APR methods. When coupled with LM-Refine post-processing, our method out-performs SOTA on outdoor scenes and under-performs SOTA on indoor scenes.

\textbf{Acknowledgement} This work was supported by the National Natural Science Foundation of China under Grant 62176010.

% \vspace{-1.3mm}

\clearpage
\bibliographystyle{splncs04}
\bibliography{egbib}
\end{document}

% \linenumbers
\pagestyle{headings}
\mainmatter
\def\ECCVSubNumber{3498}  % Insert your submission number here

\title{Supplementary Material of \\
SC-wLS: Towards Interpretable Feed-forward Camera Re-localization}

\makeatletter
\newcommand{\printfnsymbol}[1]{%
	\textsuperscript{\@fnsymbol{#1}}%
}
\makeatother
%\begin{comment}
\titlerunning{SC-wLS: Towards Interpretable Feed-forward Camera Re-localization} 
\authorrunning{X. Wu et al.} 
\author{Xin Wu\inst{1,2}\thanks{equal contribution} \and
	Hao Zhao\inst{1,3}\printfnsymbol{1} \and
	Shunkai Li\inst{4} \and
	Yingdian Cao\inst{1,2} \and
	Hongbin Zha\inst{1,2} }
\institute{Key Laboratory of Machine Perception (MOE), School of EECS, Peking University \and
	PKU-SenseTime Machine Vision Joint Lab \and
	Intel Labs China \and Kuaishou Technology \\
	\email{\{wuxin1998,zhao-hao,lishunkai,yingdianc\}@pku.edu.cn, zha@cis.pku.edu.cn} \\}
%\end{comment}
%******************

% \begin{comment}
% % If the paper title is too long for the running head, you can set
% % an abbreviated paper title here
% \end{comment}
%******************
\maketitle

% \twocolumn[{%
% \renewcommand\twocolumn[1][]{#1}%
% \maketitle
% \begin{center}
%     \centering
%     \captionsetup{type=figure}
%     \includegraphics[width=.8\textwidth,height=5cm]{8.pdf}
%     \captionof{figure}{After self-supervised fine-tuning during test time, the quality network selects better scene coordinates.}
%     \label{fig:better}
% \end{center}%
% }]

\begin{figure*}[htb]
	\begin{center}
		\includegraphics[width=\linewidth]{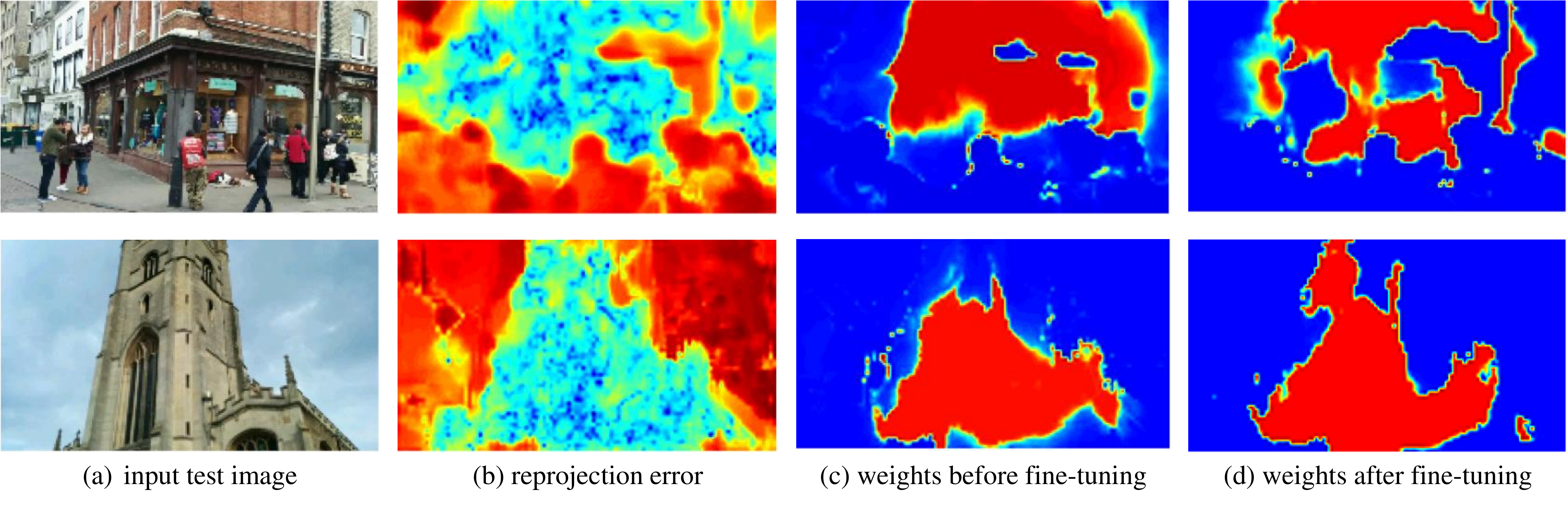}
	\end{center}
	\vspace{0pt}
	\caption{After self-supervised fine-tuning during test time, the quality network selects better scene coordinates.}
	\vspace{0pt}
	\label{fig:better}
\end{figure*}

% \maketitle

%%%%%%%%% BODY TEXT
\section{Evaluations with SfM ground truth on 7Scenes dataset}
As illustrated in \cite{brachmann2021limits}, the reference algorithm used to create pseudo ground truth has an influence on the performance of a certain family of re-localisation methods. Thus we also train (w/o 3D model) and evaluate our method using the pseudo ground truth (pGT) generated by SfM \cite{brachmann2021limits} on 7Scenes dataset. In Table~\ref{pgt7}, we report the median translational and rotational errors and recalls with pose error below 5cm, 5deg, and the settings are the same as Table 1, 4 in the main paper. Compared with the results on the original 7Scenes dataset, the performance of our methods trained with pGT-SfM improves on all of the scenes, which is consistent with other re-localization methods reported in \cite{brachmann2021limits}. Besides, it is shown that the recalls of our method Ours ($dlt$+$e2e$+$ref$) under-perform DSAC* (w/ model), and using DSAC*'s exact post-processing can compensate for this gap.

\begin{table*}[htb]

	\centering
% 	\small
% \vspace{-20pt}
	\caption{Results on the 7Scenes dataset \cite{shotton2013scene}, with translational and rotational errors measured in $m$ and $^{\circ}$, and recalls (\%) with pose error below 5$cm$, 5$^{\circ}$. Here notations are the same as the Table 1, 4 in the main paper. Note DSAC* \cite{brachmann2021limits} w/ model is trained with 3D model, and all of these methods are trained using pseudo ground truth generated by SfM \cite{brachmann2021limits}.}
	%\begin{tabular}{p{0.47cm}|p{0.47cm}|p{0.47cm}|cp{0.47cm}|cp{0.47cm}|cp{0.47cm}|cp{0.47cm}|cp{0.47cm}|cp{0.47cm}|cp{0.47cm}|cp{0.47cm}|cp{0.47cm}|cp{0.47cm}|cp{0.47cm}|cp{0.47cm}|cp{0.47cm}|cp{0.47cm}|cp{0.47cm}|cp{0.47cm}}
	 \resizebox{\textwidth}{!}{
	 
	\begin{tabular}{lcccccccc}
		%\begin{tabular}{cm|cm|cm|cm|cm|cm|cm|cm|cm|cm|cm|cm|cm}
		\toprule
		Median Errors & Chess & Fire & Heads & Office & Pumpkin & Kitchen & Stairs & \\
		\midrule
	    Ours ($dlt$) & 0.012/0.39 & 0.030/0.59 & 0.024/1.53 & 0.036/0.65 & 0.038/0.47 & 0.039/0.73 & 0.155/2.85 & \\

		Ours ($dlt$+$e2e$) & 0.011/0.37 & 0.029/0.58 & 0.022/1.46 & 0.029/0.59& 0.035/0.43 & 0.033/0.60 & 0.093/2.03 & \\ 
		Ours ($dlt$+$e2e$+$ref$) & \textbf{0.007}/\textbf{0.20} & \textbf{0.010}/\textbf{0.34} & \textbf{0.009}/\textbf{0.53} & \textbf{0.013}/\textbf{0.28} & \textbf{0.017}/\textbf{0.32} & \textbf{0.010}/\textbf{0.21} & \textbf{0.032}/\textbf{0.81} & \\
		\midrule
		Recalls & Chess & Fire & Heads & Office & Pumpkin & Kitchen & Stairs & Avg \\
		\midrule
	
		DSAC* \cite{brachmann2021limits} w/ model & 99.9 & 98.9 & \textbf{99.8} & 98.1 & \textbf{99.0} & 97.0 & \textbf{92.0} & \textbf{97.8} \\
		\midrule
		Ours ($dlt$+$e2e$) & 90.1 & 66.0 & 60.7 & 75.6 & 63.2 & 64.0 & 28.8 & 64.1  \\
		Ours ($dlt$+$e2e$+$ref$) & 100 & 86.3 & 68.9 & 93.9 & 80.5 & 91.1 & 65.9 & 83.8 \\
		Ours ($dlt$+$e2e$+$dsac*$) & \textbf{100} & \textbf{99.7} & 97.9 & \textbf{99.5} & 94.4 & \textbf{97.7} & 84.2 & 96.0 \\
	
		\bottomrule
	\end{tabular}}
	%\setlength{\abovecaptionskip}{-10cm}
	%\setlength{\belowcaptionskip}{-10cm}
% 	\vspace{7pt}
	\label{pgt7}
	\vspace{0pt}
\end{table*}

\section{More ablation studies}

% \noindent \textbf{Graph Attention Layer} 
\paragraph{Graph Attention Layer.}
As described in Section 3.2 in the main paper, we use Graph Attention Layers in the quality weight network. The quality weight network is based on OANet \cite{zhang2019learning}, which is a PointNet-like architecture that takes 2D-3D correspondence pairs as input. We propose to introduce
the self-attention mechanism into OANet, replacing the original normalized MLP modules with Graph Attention Layers. This layer builds a fully connected graph upon clusters and conducts global message passing between all nodes. This network architecture enhancement improves the re-localization accurary. As shown in Table~\ref{ablation}, using the Graph Attention Layer, the median transition error for Ours ($dlt$) is decreased from 23cm to 18cm on \emph{Stairs}, and 19cm to 15cm on \emph{Shop Facade}. We also report the results of Ours ($dlt$+$e2e$) in Table~\ref{ablation}, which are consistent with Ours ($dlt$) in most scenes.

\paragraph{Classification loss $\lossh_{c}$} in Section 3.3 is a binary cross-entropy function % (0 for outliers, 1 for inliers)
which plays the role of a hard outlier pruner. It allows a more stable convergence during training, and enhances the pose estimation accuracy. $\lossh_{c}$ is effective for both indoor and outdoor scenes, and especially useful for chanllenge environments, as shown in Table~\ref{ablation}. For instance, it results in a relative error reduction of 77\% on \emph{Shop Facade}, whose scene coordinates are very noisy due to dynamic objects and non-Lambertian reflection. 
% enhances the convergence stability and the predicted pose accuracy.

\begin{table*}[htb]
	\centering
	%\begin{tabular}{p{0.47cm}|p{0.47cm}|p{0.47cm}|cp{0.47cm}|cp{0.47cm}|cp{0.47cm}|cp{0.47cm}|cp{0.47cm}|cp{0.47cm}|cp{0.47cm}|cp{0.47cm}|cp{0.47cm}|cp{0.47cm}|cp{0.47cm}|cp{0.47cm}|cp{0.47cm}|cp{0.47cm}|cp{0.47cm}|cp{0.47cm}}
% 	\renewcommand\tabcolsep{5.0pt}
% 	\begin{threeparttable}
    \caption{More ablation studies on some scenes of 7Scenes dataset \cite{shotton2013scene} and Cambridge dataset \cite{kendall2015posenet}. We validate the effects of Graph Attention Layer and classification loss $\lossh_{c}$, and report median translational and rotational 
	errors measured in $m$ and $^{\circ}$. $dlt$ and $e2e$ are evaluated without and with the third end-to-end fine-tuning stage mentioned in Section 3.4 of the main paper. The best results are highlighted.
	}
	 \resizebox{0.85\textwidth}{!}{
	 
	    \begin{tabular}{cc|ccccc}
		%\begin{tabular}{cm|cm|cm|cm|cm|cm|cm|cm|cm|cm|cm|cm|cm}
% 		\toprule
        \hline
         & & \multicolumn{5}{c}{Ours ($dlt$)} \\
        \hline
		Attention Layer & $\lossh_{c}$  & Fire & Office & Stairs & Shop Facade & Church\\
% 		\midrule
        \hline

		\checkmark &  & 0.055/1.06 & 0.076/1.06 & 0.251/4.16 & 0.65/2.5 & 0.62/1.9 \\
		 & \checkmark & 0.060/1.09 & 0.068/1.03 & 0.230/4.00 & 0.19/1.2 & 0.50/1.5 \\
	
		 \checkmark & \checkmark & \textbf{0.051}/\textbf{1.04} & \textbf{0.063}/\textbf{0.93} & \textbf{0.179}/\textbf{3.61} & \textbf{0.15}/\textbf{1.1} & \textbf{0.50}/\textbf{1.5} \\

		\hline
         & & \multicolumn{5}{c}{Ours ($dlt$ + $e2e$)} \\
        \hline
         & \checkmark & 0.057/1.10 & 0.061/\textbf{0.84} & 0.163/3.06 & 0.12/0.7 & \textbf{0.35}/1.3 \\
	
		 \checkmark & \checkmark & \textbf{0.048}/\textbf{1.09} & \textbf{0.055}/0.86 & \textbf{0.123}/\textbf{2.80} & \textbf{0.11}/\textbf{0.7} & 0.39/\textbf{1.3} \\
        \hline
% 		\bottomrule
	\end{tabular}}
	%\setlength{\abovecaptionskip}{-10cm}
	%\setlength{\belowcaptionskip}{-10cm}
% 	\vspace{7pt}
	
	\label{ablation}
% 	\end{threeparttable}
\end{table*}

\section{Self-supervised adaptation at test time }

\begin{table*}[htb]
	\centering
	\caption{Median errors of the pose estimation on 7Scenes dataset \cite{shotton2013scene} w.r.t. the self-supervised adaptation module. Translational and rotational errors are measured in $cm$ and $^{\circ}$. $dlt$ is the weighted DLT method, $e2e$, $ref$ and $self$ denote end-to-end training step, LM-Refine and self-supervised adaptation, respectively. 150k means 150k iterations of fine-tuning. The best results are highlighted.}
	\resizebox{\textwidth}{!}{%
	
% 	\resizebox{2\columnwidth}{!}{
	%\begin{tabular}{p{0.47cm}|p{0.47cm}|p{0.47cm}|cp{0.47cm}|cp{0.47cm}|cp{0.47cm}|cp{0.47cm}|cp{0.47cm}|cp{0.47cm}|cp{0.47cm}|cp{0.47cm}|cp{0.47cm}|cp{0.47cm}|cp{0.47cm}|cp{0.47cm}|cp{0.47cm}|cp{0.47cm}|cp{0.47cm}|cp{0.47cm}}
	\begin{tabular}{lccccccc}
		%\begin{tabular}{cm|cm|cm|cm|cm|cm|cm|cm|cm|cm|cm|cm|cm}
		\toprule
		Methods & Chess & Fire & Heads & Office & Pumpkin & Kitchen & Stairs\\
		\midrule

		Ours ($dlt$+$e2e$+$self$ [150k])& 2.5/0.71 & 3.3,0.98 & 2.6/1.84 & 4.1/0.81 & 6.1/1.21 & 6.2/1.26 & 6.8/1.24 \\
		Ours ($dlt$+$e2e$+$self$ [600k])& 2.1/0.64 & \textbf{2.3}/\textbf{0.80} & 1.3/0.79 & 3.7/0.76 & \textbf{4.8}/1.06 & \textbf{5.1}/\textbf{1.08} & 5.5/1.48 \\
% 		Ours($dlt$+$e2e$+$self$+$ref$ [150k])& 1.9/0.63 & 2.5/0.87 & 1.4/0.86 & 3.7/0.80 & 5.4/1.12 & 5.6/1.17 & 5.0/1.26 \\
		Ours ($dlt$+$e2e$+$self$+$ref$ [600k])& \textbf{1.9}/\textbf{0.62} & 2.4/0.86 & \textbf{1.3}/\textbf{0.69} & \textbf{3.4}/\textbf{0.76} & 4.9/\textbf{1.04} & 5.2/1.12 & \textbf{4.7}/\textbf{1.21} \\
		%ours w/ 50\% RANSAC & 0.02/1.71 & 0.035/1.13 & 0.018/1.02 & 0.04/0.94 & & & 0.06/1.54\\
% 		ours w/ LM-refine & 0.019/0.64 & 0.027/0.90 & 0.0149/0.88 & & 0.046/1.07 & & 0.07/2.04\\
% 	    \midrule
% 		Ours($dlt$+$e2e$+$ref$)& \textbf{1.9}/\textbf{0.62} & 2.5/0.88 & 1.4/0.9 & 3.5/0.78 & 5.1/1.07 & 5.4/1.15 & 6.3/1.28 \\
		
		\bottomrule
	\end{tabular}}
	%\setlength{\abovecaptionskip}{-10cm}
	%\setlength{\belowcaptionskip}{-10cm}
% 	\vspace{7pt}
% 	}
	\label{self7}
\end{table*}

As illustrated in the main paper, self-supervised adaptation deals with the domain shift problem during test time and significantly enhances the re-localization performance of the DLT setting. As a supplement to Fig. 4 in the main paper, we compare learned weights on Cambridge before and after self-supervised adaptation at test-time, in Fig.~\ref{fig:better}. It shows that the quality network is capable of selecting better scene coordinates after test-time adaptation.

% On one hand, we want to know whether the network can adapt with limited iterations so that it is practical in real-world applications. On the other hand, we want to know how accurate this adaptation can be if unlimited iterations are allowed. 
\begin{table*}[htb]
	\centering
	\caption{Median errors of the pose estimation on Cambridge dataset \cite{kendall2015posenet} w.r.t. the self-supervised adaptation module. Translational and rotational errors are measured in $m$ and $^{\circ}$. The notations are the same as Table~\ref{self7}. The best results are highlighted.
	}
	\resizebox{\textwidth}{!}{%
	
	%\begin{tabular}{p{0.47cm}|p{0.47cm}|p{0.47cm}|cp{0.47cm}|cp{0.47cm}|cp{0.47cm}|cp{0.47cm}|cp{0.47cm}|cp{0.47cm}|cp{0.47cm}|cp{0.47cm}|cp{0.47cm}|cp{0.47cm}|cp{0.47cm}|cp{0.47cm}|cp{0.47cm}|cp{0.47cm}|cp{0.47cm}|cp{0.47cm}}
	\begin{tabular}{lccccc}
		%\begin{tabular}{cm|cm|cm|cm|cm|cm|cm|cm|cm|cm|cm|cm|cm}
		\toprule
		Methods & Greatcourt & King's College & Shop Facade & Old Hospital & Church\\
		\midrule

		Ours ($dlt$+$e2e$+$self$ [150k]) & 0.95/0.5 & 0.11/0.4 & 0.05/0.4 & 0.20/0.7 & 0.22/0.9 \\
		Ours ($dlt$+$e2e$+$self$ [600k]) & 0.94/0.5 & 0.11/0.3 & 0.05/0.4 & 0.18/0.7 & 0.17/0.8 \\
	
% 		Ours($dlt$+$e2e$+$self$+$ref$ [150k]) & 1.97/1.8 & 0.08/0.2 & 0.05/0.4 & 0.10/0.4 & 0.12/0.5\\
		Ours ($dlt$+$e2e$+$self$+$ref$ [600k]) & \textbf{0.28}/\textbf{0.2} & \textbf{0.08}/\textbf{0.2} & \textbf{0.04}/\textbf{0.3} & \textbf{0.11}/\textbf{0.4} & \textbf{0.09}/\textbf{0.3} \\
% 		\midrule
		%ours w/ 50\% RANSAC & 0.02/1.71 & 0.035/1.13 & 0.018/1.02 & 0.04/0.94 & & & 0.06/1.54\\
% 		Ours($dlt$+$e2e$+$ref$) & 9.44/6.6 & \textbf{0.08}/0.3 & \textbf{0.05}/\textbf{0.4} & 0.11/0.4 & 0.13/0.5 \\
        % Ours ($dlt$+$e2e$+$ref$)  & 0.29/\textbf{0.2} & \textbf{0.08}/\textbf{0.2} & \textbf{0.04}/\textbf{0.3} & \textbf{0.11}/\textbf{0.4} & \textbf{0.09}/\textbf{0.3} \\

		\bottomrule
	\end{tabular}}
	%\setlength{\abovecaptionskip}{-10cm}
	%\setlength{\belowcaptionskip}{-10cm}
% 	\vspace{7pt}
	\label{self_cam}
\end{table*}

\begin{figure}[htb]
	\begin{center}
		\includegraphics[scale=0.45]{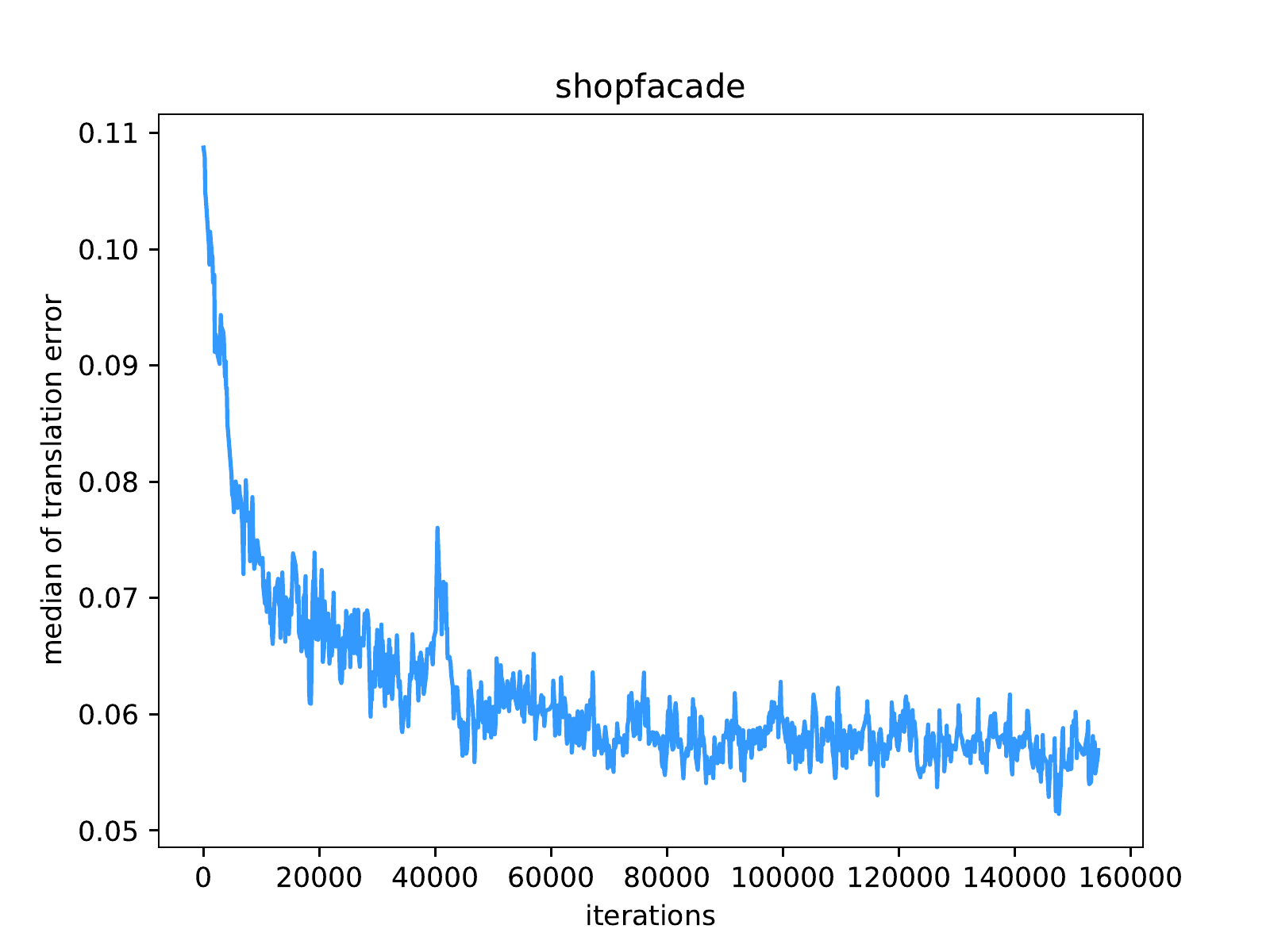}
	\end{center}
	\caption{The median translation error of the estimated poses on \emph{shopfacade} (from Cambridge) during self-supervised adaptation. It rapidly converges within 20k iterations, which demonstrates the effectiveness of this module in real world scenarios.}
	\label{fig:errdrop}
\end{figure}

In the main paper, we report the results of fine-tuning for 600k iterations on 7Scenes and Cambridge datasets. It is important to investigate how well the network adapt with fewer iterations, so as to be practical in real-world applications. As shown in Fig.~\ref{fig:errdrop}, the median translation error of the estimated poses drops rapidly within 20k iterations (wall-clock time: 23 minutes). Thus, we show additional pose estimation results within fewer iterations (150k, wall-clock time: 2.9 hours), in Table~\ref{self7} and \ref{self_cam}. It demonstrates that the proposed self-supervised adaptation module can achieve reasonably good results using few iterations, and fine-tuning with more iterations proceeds to improve the pose estimation.

\begin{figure*}[htb]
	\begin{center}
		\includegraphics[width=\linewidth]{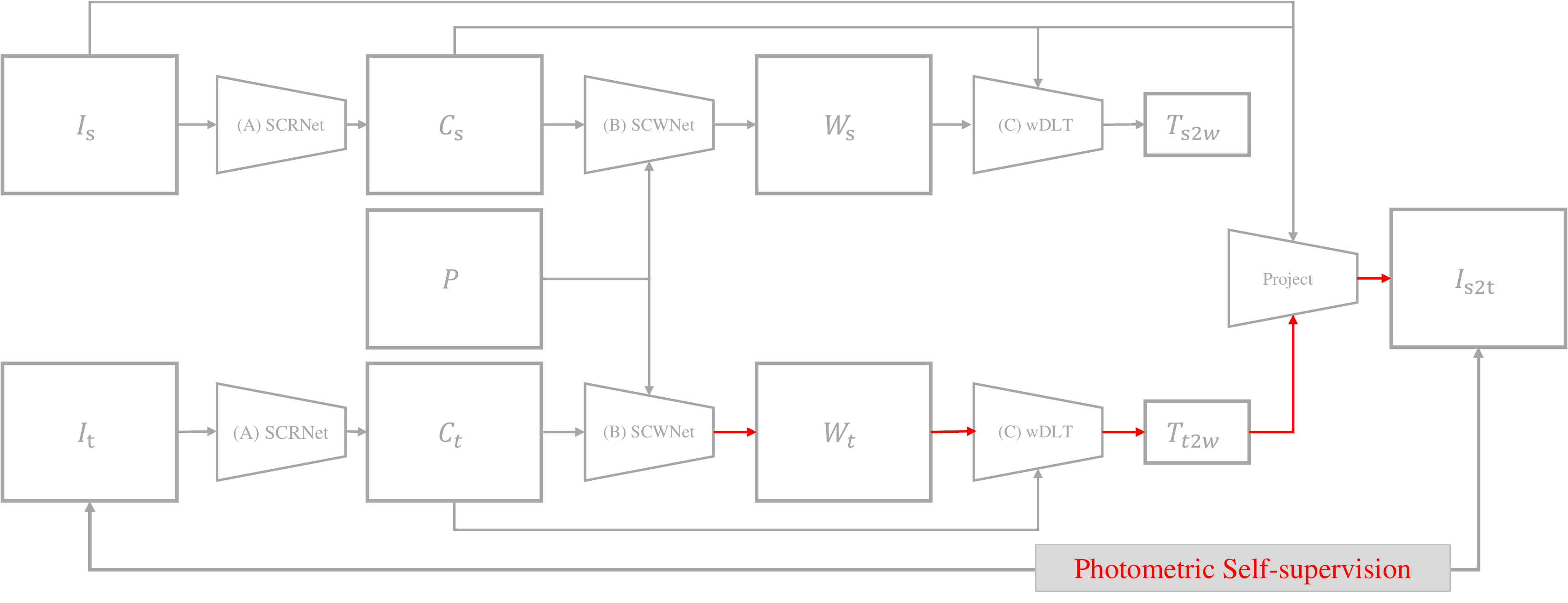}
	\end{center}
	\caption{A detailed illustration of the self-supervised fine-tuning procedure. $I_s$ is the source RGB image. $I_t$ is the target RGB image. SCRNet (A) is the scene coordinate regression network. $C_s$ is the scene coordinate map for the source image. $C_t$ is the scene coordinate map for the target image. $P$ is the 2D pixel coordinate map. SCWNet (B) is the scene coordinate quality weight network. $W_s$ is the scene coordinate weight map for the source image. $W_t$ is the scene coordinate weight map for the target image. wDLT (C) is the weighted least squares solver. $T_{s2w}$ is the camera pose of the source image in the world frame. $T_{t2w}$ is the camera pose of the target image in the world frame. Only tensors that flow through red arrows receive supervision signals while others are detached.}
	\label{fig:ftback}
\end{figure*}

Besides, we describe the mechanism of self-supervised adaptation in detail, using Fig.~\ref{fig:ftback}. Specifically, we select two adjacent frames as source and target images, and warp $\img_{s}$ to target $\img_{t}$ using the predicted scene coordinates $\mathbf{C_{s}}$ in the world coordinate system and the transform matrix $\mathbf{T_{t2w}}$ calculated by the DLT process. Then the photometric error serves as self-supervision and back-propagates gradients along the red arrows, while others are detached in this stage. Note that the $\mathbf{T_{t2w}}$ is post-processed by the algorithm of Section~\ref{sec:postDLT}, which is fully differentiable and enables the gradients flowing from the photometric loss. In our experiments, the sampling interval of two images is 7 for 7-Scenes and 1 for Cambridge, respectively.

\section{3D Map visualization}
In Fig.~\ref{fig:map}, we visualize point clouds predicted directly by scene coordinate regression networks and those filtered by the learned weights of our SC-wLS framework. We filter out points with weights smaller than $\lambda=0.9$. It shows that the scene coordinates are particularly noisy and due to the nature of reprojection supervision, point clouds may drift away far along the bearing vectors. However, with the assistance of our SC-wLS, the weight-filtered point clouds are much more accurate and cleaner, showing the effectiveness and interpretability of the learned weights as well.

\section{Loss functions}
This section elaborates on the hyper-parameter tuning of the regression loss $\lossh_{r}$ in Section 3.3. As illustrated in the main paper, we use an eigen-decomposition free loss \cite{Dang2020EigendecompositionFreeTO} to refrain from the numerical instability caused by the eigen-vector switching problem:
\begin{equation}
    \begin{aligned}
      \lossh_{r} =
     \be^\top \bX^\top \rm diag(\wei) \bX \be\ + \alpha e^{-\beta tr(\bbX^\top \rm diag(\wei) \bbX)} \,
    \end{aligned}
    \label{eq:ell_comb}
  \end{equation}
  where $\be$ is the flattened ground-truth pose, $\bbX {=} \bX(\bI - \be\be^\top)$, while $\alpha$ and $\beta$ are positive scalars.
  
As mentioned before, those two terms in Eq.~\ref{eq:ell_comb} serve as different roles. The former originates from $\rm diag(\sqrt{\wei})\bX \be\ = 0$, and minimizing this term leads to the trivial solution of $\wei=\textbf{0}$ as well. Thus the latter term is proposed to alleviate its impact. Since $\bbX$ projects all data vectors onto the hyperplane normal to $\be$, we could maximize the trace of $\bbX^\top \rm diag(\wei) \bbX$ to make the eigenvalues corresponding to directions orthogonal to $\be$ to be as large as possible. It's important to select proper hyperparameters $\alpha$ and $\beta$ to balance these two terms. At the same time, the magnitude of the last term varies considerably on different scenes. We recommend to choose $\beta$ as the inverse of the magnitude of this trace, and in our experiments, $\alpha$ and $\beta$ are set to 5 and 1e-4 for indoor 7Scenes while 5 and 1e-6 for outdoor Cambridge, respectively. %Moreover, we observe that in the late stage of training, slightly decreasing the value of $\alpha$ can further enhance the pose estimation performance in some degrees, which is rational since the first term pays more attention to the accuracy of learning.

\section{Network architecture}

As mentioned in Section 4.1 in the main paper, we adopt the scene coordinate regression network architecture from \cite{brachmann2021visual} for 7Scenes. For Cambridge,
% we additionally extend the early layers of this network and use residual connections, as shown in Fig ?. 
we use the same architecture as the feature encoder in RAFT \cite{teed2020raft} to replace the early layers of this network, which has residual connections, as shown in Fig~\ref{fig:network}. 
It increases the receptive field size of the network, and enhances the robustness of feature extraction for complicated environments. We have also tried this enhanced architecture for 7Scenes, which does not bring performance improvements.

\section{Post-processing of the DLT algorithm}
\label{sec:postDLT}
As described in Section 3.1 of the main paper, pose estimation is solved by the Direct Linear Transform (DLT) algorithm:
\begin{equation} \label{eq:LS}
    \mathbf X^{\top} \rm diag(\wei)\mathbf X  \rm Vec(\tr) = 0
\end{equation}
where the transform matrix is flattened as $\rm Vec(\tr)$, and corresponds to the smallest eigen-vector of $\mathbf X^{\top} \rm diag(\wei)\mathbf X$. To guarantee that the rotation matrix $\rotm$ is orthogonal and has determinant 1, we post-process the DLT results using the common generalized Procrustes algorithm \cite{schonemann1966generalized}. The pseudo-code of our implementation is:

% Note that in SC-wLS, each correspondence contributes differently according to $\wei_{i}$, so $\mathbf X^{T}\mathbf X$ can be rewritten as $\mathbf X^{T}diag(\wei)\mathbf X$
% and $Vec(\tr)$ corresponds to its smallest eigenvector.
% As the rotation matrix $\rotm$ needs to be orthogonal and has determinant 1, we futher refine the DLT results by the generalized Procrustes algorithm \cite{schonemann1966generalized}, consisting of SVD decomposition, which is also differentiable. More details can be found in the supplementary material.

\begin{algorithm}[htb]  
  \caption{Post-processing of the pose calculated by the DLT algorithm}  
  \begin{algorithmic}[1]
    \State $ \textbf{Input:} \bar{\tr} =  \begin{bmatrix}
 p_{1} &p_{2} &p_{3} &p_{4} \\ 
 p_{5} &p_{6} &p_{7} &p_{8} \\ 
 p_{9} &p_{10} &p_{11} &p_{12} \\
\end{bmatrix}=[\bar{\rotm}_{3\times 3} | \bar{\trans}_{3\times 1}]$, and learned weight $\wei$;
    \State \textbf{Output:} Regularized $\tr = [\rotm_{3\times 3} | \trans_{3\times 1}]$;
    %\State $\bar{\rotm}, \bar{\trans} = \bar{\tr}$;
    \State $\mathbf{U} \mathbf{\Sigma} \mathbf{V} = \rm SVD(\bar{\rotm})$;
    \State $s = \frac{3}{tr(\mathbf{\Sigma})} $;
    
    \State Selecting the most confident 2D-3D correspondence $c = (u,v,x,y,z)$ according to the learned weight $\wei$;
    
    \If{$s(xp_{9} + yp_{10} + zp_{11} + p_{12}) > 0$}
    \State $s = s$;
    \Else
    \State $s = -s$;
    \EndIf
    \State $\rotm = sign(s) \mathbf{U} \mathbf{V}^{\top}$;
    \State $ \trans  = s\bar{\trans}$;
    \label{code:recentEnd}  
  \end{algorithmic}  
\end{algorithm}  
It's worth noting that these steps are fully differentiable, thus the self-supervised adaptation in Section 3.6 is able to back-propagate gradients through this operation. 

\textbf{
\begin{figure*}[htb]
	\begin{center}
		\includegraphics[width=\linewidth]{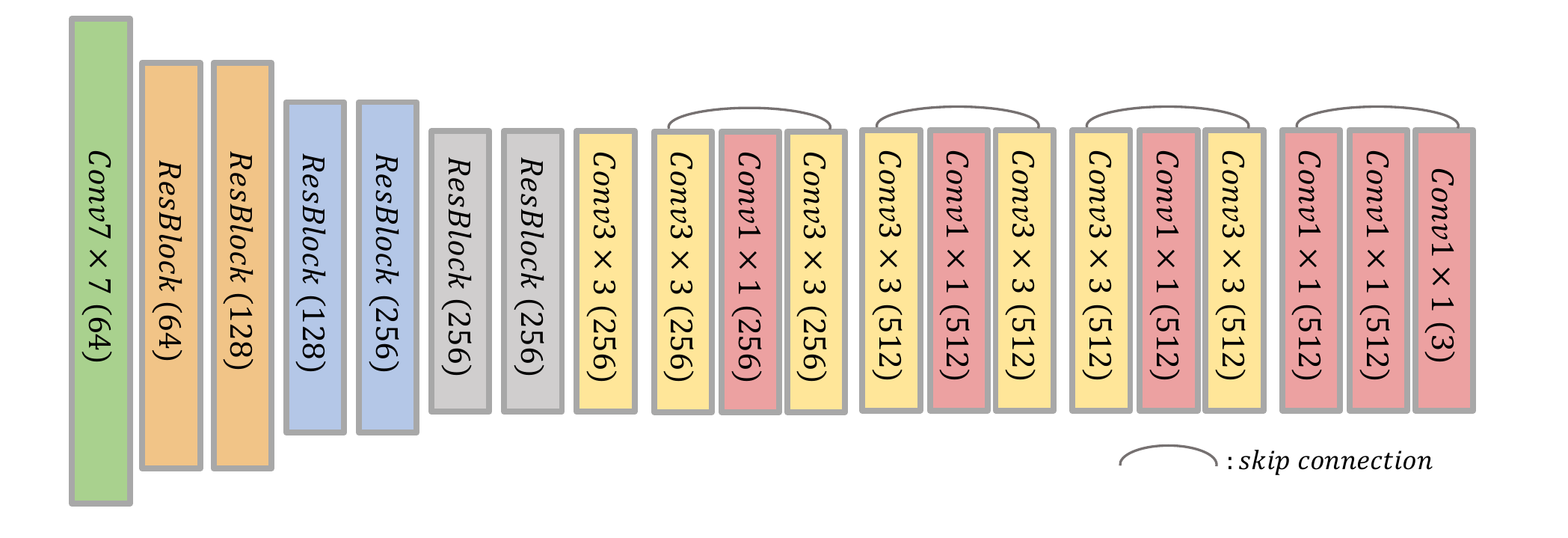}
	\end{center}
	\caption{Scene coordinate regression network details for Cambridge. The first 8 blocks are the same as the feature encoder in \cite{teed2020raft}, and no normalization is used. Skip connections are used in both ResBlocks and the last 12 blocks as denoted.}
	\label{fig:network}
\end{figure*}}

\begin{figure*}[htb]
	\begin{center}
		\includegraphics[width=\linewidth]{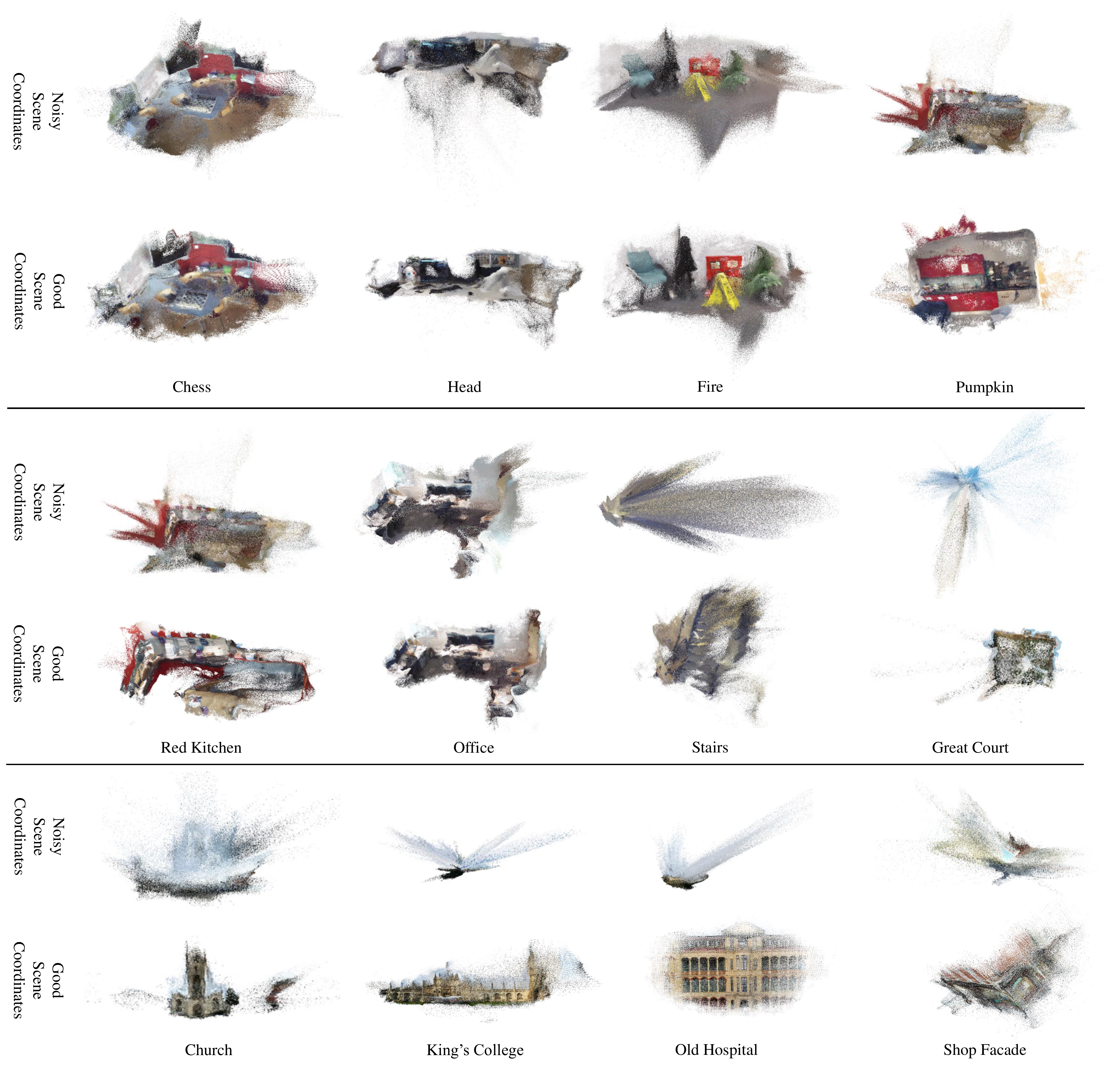}
	\end{center}
	\caption{Map visualization on the test sets of 7-Scenes and Cambridge. Since scene coordinates are predicted in the world frame, we directly show the point clouds generated by aggregating scene coordinate predictions on test frames. It can be seen that only predicting scene coordinates results in noisy point clouds, especially in outdoor scenes where scene coordinate predictions on sky regions are only meaningful in term of their 2D projections. We show good scene coordinates by filtering out samples with a quality weight lower than 0.9.}
	\label{fig:map}
\end{figure*}

\section{Error Metrics}
\begin{figure}[htb]
	\centering
	%\fbox{\rule{0pt}{0.5in} \rule{0.9\linewidth}{0pt}}
	%	\includegraphics[width=\linewidth]{err_per_frame.pdf}
	\includegraphics[width=0.9\linewidth]{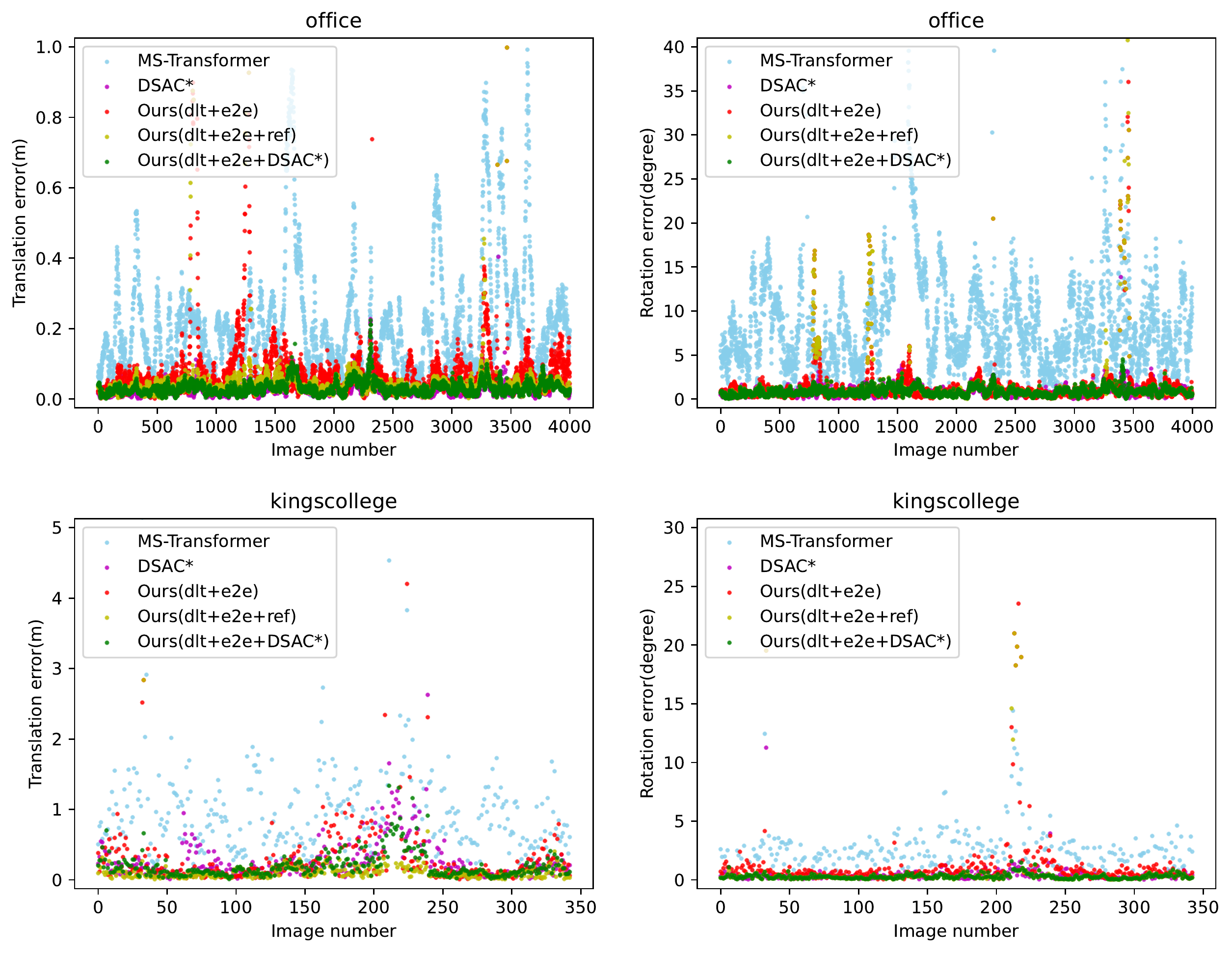}
	\vspace{0pt}
	\caption{Translation and rotation errors per testing frame.}
	%	 on \emph{office} and \emph{kingscollege}
	\vspace{0pt}
	\label{fig:perframe}
\end{figure}

Median errors are robust to outlier estimates. We report the translation and rotation errors per testing frame on \emph{office} and \emph{kingscollege} in Fig.~\ref{fig:perframe}. It's shown that per frame error may be negatively impacted by outliers thus hard to tell the true algorithm performances, while median errors are easier to compare between lots of methods. We have released the code for per frame error analysis.  

\section{Delving into Interpretability}

As for interpretability, we report pearson correlation coefficients between learned weights and inverse reprojection errors in Table~\ref{tab:pearson}, in which outdoor scenes give higher correlation values due to many uncertain regions like sky and human.

\begin{table}[htb]
	\centering
	\caption{Pearson correlation coefficients. ($p < 0.01$)}
	%\small
	% \footnotesize
	%\begin{tabular}{p{0.47cm}|p{0.47cm}|p{0.47cm}|cp{0.47cm}|cp{0.47cm}|cp{0.47cm}|cp{0.47cm}|cp{0.47cm}|cp{0.47cm}|cp{0.47cm}|cp{0.47cm}|cp{0.47cm}|cp{0.47cm}|cp{0.47cm}|cp{0.47cm}|cp{0.47cm}|cp{0.47cm}|cp{0.47cm}|cp{0.47cm}}
	%\resizebox{0.5\textwidth}{!}{
		%	\setlength{\tabcolsep}{0.1mm}
		\begin{tabular}{c|c|c|c|c|c}
			%\begin{tabular}{cm|cm|cm|cm|cm|cm|cm|cm|cm|cm|cm|cm|cm}
			%\toprule
			\hline
			Chess & Fire & Heads & Office & Pumpkin & Kitchen \\
			\hline
			0.18 & 0.15 & 0.18 & 0.15 & 0.21 & 0.14 \\
			\hline
			Stairs & Court & College & Shop & Hospital & Church  \\
			%	\midrule
			\hline
			0.21 & 0.23 & 0.22 & 0.23 & 0.24 & 0.21 \\
			\hline
			%\bottomrule
	\end{tabular}
	\label{tab:pearson}
\end{table}

%
%NG-SAC like methods do this way, but it learns weights from RGB images wchih can be troubled by domain biases. While our weights are learned from 5D point clouds that capture the geometric structure, which is not impacted by photometric variations.
%
%it introduces
%geometric inductive biases and decomposes the system
%into two black boxes of different functions: (1) the first
%one predicts explicit 2D-3D correspondences; (2) the second
%one learns geometric patterns from 5D point clouds, facilitated
%by differentiable least squares.
%
%direct end-to-end training from scratch diverges. 
 
%
%Firstly, the effectiveness of $L_c$ is valiated by this experiemtn (...) Secondly, this loss is inspired by the Equation 7 of Learning to find good correspondences CVPR 2018.
%This has already been shown in supp and we have added a link in the main paper.

\bibliographystyle{splncs04}
\bibliography{supple_bib}